\documentclass[runningheads]{llncs}

\usepackage[table,usenames,dvipsnames]{xcolor}

\usepackage{eccv}

\usepackage{eccvabbrv}

\usepackage{array,booktabs}
\usepackage{caption}
\newcolumntype{M}[1]{>{\centering\arraybackslash}m{#1}} %
\usepackage{subcaption} %
\usepackage{graphicx}

\usepackage{booktabs}
\usepackage{amsmath}
\usepackage{tabularx}
\usepackage{wrapfig}
\usepackage{comment}

\usepackage{afterpage}

\renewcommand{\figurename}{Figure}

\usepackage{hyperref}

\usepackage{orcidlink}

\newcommand{\subject}{\texttt{<subject>}}
\newcommand{\object}{\texttt{<object>}}
\newcommand{\predicate}{\texttt{<predicate>}}
\newcommand{\triplet}{\texttt{<subject-predicate-object>}}
\newcommand{\tuple}{\texttt{<subject-object>}}

\newcolumntype{C}{>{\centering\arraybackslash}X}

\usepackage{letltxmacro}
\LetLtxMacro{\oldsubsubsection}{\subsubsection}
\renewcommand{\subsubsection}[1]{\vspace{-4mm}\oldsubsubsection{#1}}

\newcommand\blfootnote[1]{%
  \begingroup
  \renewcommand\thefootnote{}\footnote{#1}%
  \addtocounter{footnote}{-1}%
  \endgroup
}

\begin{document}

\title{Scene-Graph ViT: End-to-End Open-Vocabulary Visual Relationship Detection} 

\titlerunning{Scene-Graph ViT}

\author{Tim Salzmann$^\dagger$\inst{1,2}\orcidlink{0000-0002-7305-2667} \and
Markus Ryll \inst{2}\orcidlink{0000-0003-2203-2946} \and \\
Alex Bewley$^\star$ \inst{1}\orcidlink{0000-0002-8428-9264} \and
Matthias Minderer$^\star$ \inst{1}\orcidlink{0000-0002-6428-8256}
}

\authorrunning{T.~Salzmann et al.}

\institute{\hspace{-0.3em}Google DeepMind \\
\email{\{tsal, bewley, mjlm\}@google.com}
\and
\hspace{-0.3em}Technical University Munich \\
\email{\{tim.salzmann, markus.ryll\}@tum.de}
}

\maketitle

\blfootnote{
$^\dagger$Work done while at Google DeepMind.\\
$^\star$Advising project leads in alphabetical order.}

\begin{abstract}
\begin{figure}
    \centering
    \vspace{-12mm}
    \includegraphics[width=\linewidth]{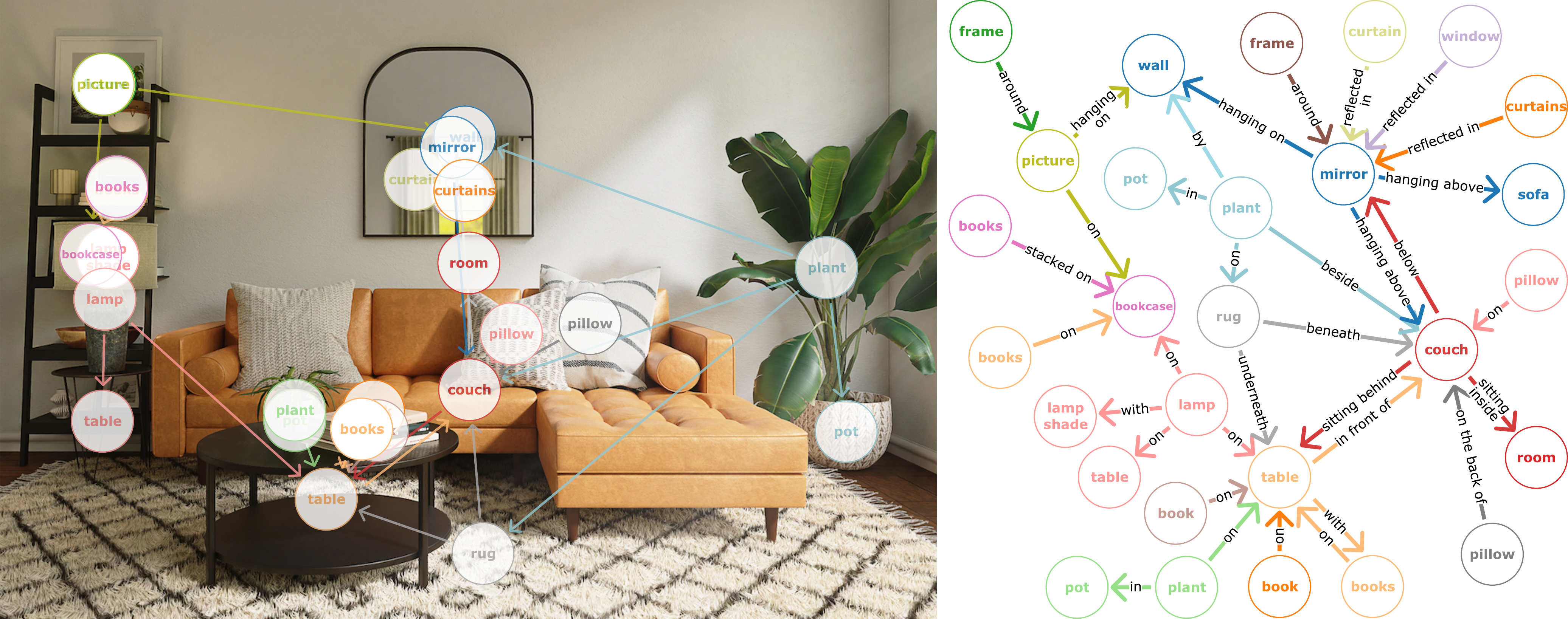}
    \vspace{-6mm}
    \caption{Relationships detected by our method on an unseen image. The top relationships by confidence score are shown. Photo by Spacejoy on \href{https://unsplash.com/photos/brown-wooden-bed-frame-with-white-and-brown-bed-linen-umAXneH4GhA}{Unsplash}.}
    \label{fig:figure_1}
\end{figure}
\vspace{-1mm}

Visual relationship detection aims to identify objects and their relationships in images. Prior methods approach this task by adding separate relationship modules or decoders to existing object detection architectures. This separation increases complexity and hinders end-to-end training, which limits performance. We propose a simple and highly efficient decoder-free architecture for open-vocabulary visual relationship detection. Our model consists of a Transformer-based image encoder that represents objects as tokens and models their relationships implicitly. To extract relationship information, we introduce an attention mechanism that selects object pairs likely to form a relationship. We provide a single-stage recipe to train this model on a mixture of object and relationship detection data. Our approach achieves state-of-the-art relationship detection performance on Visual Genome and on the large-vocabulary GQA benchmark at real-time inference speeds. We provide ablations, real-world qualitative examples, and analyses of zero-shot performance.

\end{abstract}
\vspace{3mm}

\section{Introduction}

A fundamental goal of computer vision is to decompose visual scenes into structured semantic representations. 
A commonly studied task towards this goal is object detection, in which objects in an image are localized by bounding boxes and classified into semantic categories.
However, a scene description also includes the semantic \emph{relationships} between objects. This gives rise to the task of visual relationship detection (VRD, \cite{lu2016visual, xu2017scene, zellers2018neural, chao2018, yang2018graph, chen2019knowledge}). In VRD, the model detects objects and infers pairwise relationships between them in the form of \texttt{<subject-predicate-object>} triplets\footnote{We use \texttt{<fixed width font>} to distinguish the grammatical terms \texttt{<subject>} and \texttt{<object>} from generic use of the word \emph{object} as in \emph{object detection}.}~\cite{lu2016visual}.

Detecting both objects and their relationships allows the construction of a \emph{scene graph}~\cite{Johnson_2015_CVPR, li2017scene, xu2017scene} in which objects are represented as nodes and their relationships as edges. Scene graph generation (SGG) has wide-ranging applications in robotics~\cite{rana2023sayplan, amiri2022reasoning, chen2023open, gu2023conceptgraphs, singh2023scene, hughes2022hydra} and image retrieval~\cite{Johnson_2015_CVPR, LI2024127052}. Increasingly, structured scene representations are also employed to provide grounding and explainability to multimodal large language models~\cite{herzig2023incorporating, zhang2024groundhog, peng2024grounding}.

Prior work typically draws a distinction between object detection and relationship prediction. Detection is performed either as a wholly separate step before relationship prediction~\cite{lu2016visual, xu2017scene, zellers2018neural, yu2020cogtree, zhang2022fine, desai2021learning}, or by separate model parts such as ``relationship decoders'' that are responsible for modeling the interactions between objects~\cite{univrd, cong2023reltr, yang2018graph}. This distinction makes it hard to optimize such models end-to-end for VRD~\cite{univrd}. In contrast, we propose an encoder-only architecture that models objects and relationships jointly, directly in the image encoder. Our architecture performs open-vocabulary relationship detection and can be trained end-to-end on arbitrary mixtures of object detection and relationship annotations.

Our model builds on a Transformer-based encoder-only object detector~\cite{minderer2022simple} in which the output tokens of the image encoder directly represent object proposals. From these tokens, class embeddings and bounding boxes are decoded with light-weight heads. Our insight is that this architecture is perfectly set up to learn relationships between objects directly in the image encoder, without the need for additional relationship-specific stages. This is because the existing self-attention in the encoder already models all-to-all pairwise interactions between the object proposal tokens.

To access information about the relationship between two of these tokens, we combine the embeddings corresponding to the \texttt{<subject>} and \texttt{<object>} token using a new \emph{Relationship Attention} layer.
Obtaining relationship embeddings for all possible pairwise combinations of object proposal tokens would be computationally infeasible. To reduce the number of combinations, we introduce a  self-supervised hard attention mechanism that selects the highest-confidence \texttt{<subject-object>} pairs at a computational cost comparable to a single self-attention layer.
We show how to directly supervise the attention scores of this mechanism without the need to propagate gradients through the hard selection.

An additional benefit of our design is that it disentangles object names from relationship predicates during inference. In contrast to prior open-vocabulary methods~\cite{univrd, yuan2022rlip, yuan2023rlipv2}, our model can embed object and predicate texts separately and generate confidence scores efficiently for all possible \texttt{<subject-}\hspace{0pt}\texttt{predicate-}\hspace{0pt}\texttt{object>} combinations.

In summary, we make the following contributions:
\begin{enumerate}
    \item An efficient architecture for open-vocabulary visual relationship detection.
    \item A single-stage recipe for joint object and relationship detection training.
    \item Efficient, disentangled object and relationship inference.
    \item Analysis of inference speed, ablations, and qualitative examples.
\end{enumerate}

Our method achieves state-of-the-art visual relationship detection performance on the Visual Genome dataset (26.1\% mR@100 with graph constraint) and on the large-vocabulary GQA benchmark.
It provides strong results while being significantly simpler than prior approaches to visual relationship detection.

\section{Related Work}

Below, we review the main approaches to VRD that are relevant to our work.
For a comprehensive overview of the field, we refer to~\cite{LI2024127052}.

\subsubsection{Detector-Agnostic Relationship Detection.}
Historically, many works on VRD assume that object detections are given, such that the task reduces to inferring relationships between the objects.
These ``detector-agnostic'' methods use off-the-shelf detectors such as Faster-RCNN~\cite{ren2015faster} to obtain boxes and box embeddings and infer a scene graph from them. 
Early work on large-scale VRD employed word embeddings to improve relationship generalization~\cite{lu2016visual} and graph inference methods such as message passing for SGG~\cite{xu2017scene}.
More recent methods leverage relationship co-occurrence statistics~\cite{zellers2018neural} and address the long-tailed nature of the training data~\cite{yu2020cogtree, zhang2022fine, desai2021learning}.

\subsubsection{End-to-End Relationship Detection.}
A fundamental limitation of the detec\-tor-agnostic approach is that object detection and relationship prediction are treated separately and cannot learn from each other. Overcoming this limitation has recently motivated the development of end-to-end architectures in which objects and relationships are predicted jointly by a single model~\cite{kim2021hotr, li2022sgtr, cong2023reltr, univrd}.
All of these models have in common that they use a Transformer~\cite{vaswani2017attention} decoder to predict object and relationship embeddings.
The decoder represents object proposals with \emph{query embeddings} that can cross-attend into image embeddings.
The literature on object detection suggests that this choice may be problematic, because the query embeddings can be difficult to initialize, which can lead to unstable and slow optimization~\cite{zhu2021deformable, yao2021efficient, song2022vidt, zhang2022dino, detr, minderer2022simple}.
To avoid this issue, we design an encoder-only architecture in which both object and relationship embeddings are computed directly from the output tokens of a Vision Transformer~\cite{dosovitskiy2020vit} image encoder, building on an idea from object detection~\cite{minderer2022simple}.
In contrast to prior models~\cite{univrd}, this allows us to train the model end-to-end on object and relationship annotations simultaneously.

\subsubsection{Long-Tailed and Open-Vocabulary Relationship Detection.}
Object clas\-ses in the natural world follow a long-tailed distribution~\cite{gupta2019lvis, zhou2022detecting}, and this is compounded in VRD due to the combination of subject, object, and predicate in a relationship~\cite{LI2024127052 , yang2018graph, xu2017scene}. 
A large number of VRD methods specifically aim to address this issue, for example with debiasing losses~\cite{yu2020cogtree}, data augmentation~\cite{zhang2022fine}, or data resampling~\cite{desai2021learning}.
Additionally, while VRD models historically assumed a fixed set of object and predicate classes, recent works add open-vocabulary capabilities~\cite{Tao_ECCV22, shi2023open, univrd, yuan2023rlipv2}.
These models leverage pretrained vision-language models such as CLIP~\cite{radford2021learning, jia2021scaling, yu2022coca, zhai2022lit} to inherit their natural-language understanding. Instead of learning a fixed set of classes, open-vocabulary models predict class embeddings that can be compared to the text embeddings of arbitrary object or predicate descriptions for classification.
Here, we employ a pretrained vision-language backbone and combine it with a carefully designed, lightweight relationship detection head to preserve and transfer semantic knowledge from the backbone pretraining to VRD.
This allows our model to achieve strong results on the large-vocabulary GQA benchmark~\cite{hudson2019gqa} (\Cref{sec:open-vocab-results}).

While we focus on architectural improvements, another line of work addresses the scarcity of training data for VRD. For example, RLIP~\cite{yuan2022rlip} and RLIPv2~\cite{yuan2023rlipv2}, which focus on human-object interaction~\cite{chao2018}, propose relationship-focused image-text pretraining and self-training, which yield large improvements in VRD performance. These works are orthogonal to our contributions and can be combined.

\section{Scene-Graph ViT}\label{sec:model}

\begin{figure}[t]
    \centering
    \includegraphics[width=0.85\linewidth]{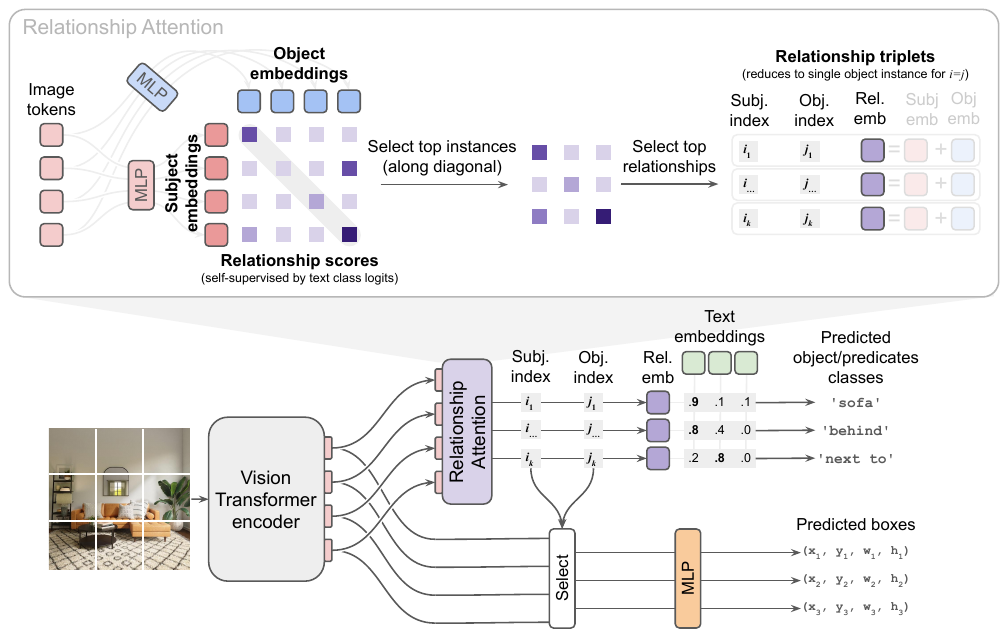}
    \caption{
    For relationship selection, image tokens are first projected using two lightweight MLPs to produce \subject{} and \object{} embeddings. A relationship score is then computed as the inner product between all \subject{} and \object{} embeddings. 
    Relationships are filtered by first selecting the top object instances, using the scores along the diagonal to represent instance likelihood. Among the remaining instances, the top \texttt{<subject-object>} pairs are selected using the off-diagonal scores. This yields a set of relationship triplets, each consisting of a \subject{} index, an \object{} index, and a relationship embedding that is computed by summing the respective \subject{} and \object{} embeddings. 
    For classification, the relationship embeddings are compared against text embeddings of object class or predicate text descriptions.
    }
    \label{fig:architecture}
\end{figure}

We propose an architecture for open-vocabulary relationship prediction in which both objects and the relationships between them are handled as ``first-class citizens'' directly in a single-stage process within the model backbone.
We build on an encoder-only architecture for object detection~\cite{minderer2022simple} and adapt it to relationship prediction by adding a specialized attention layer.
This layer exploits the pairwise structure between existing object embeddings to obtain relationship embeddings as schematized in \Cref{fig:architecture} and described below.

\subsection{Encoder-Only Open-Vocabulary Object Detection}

We briefly review the detection architecture that forms the basis of our model~\cite{minderer2022simple}.
The model consists of Transformer-based~\cite{dosovitskiy2020vit} image and text encoders that are contrastively pretrained on a large number of image and text pairs~\cite{radford2021learning}.
The image encoder is adapted to detection by removing the final pooling layer and adding heads that predict bounding boxes and class embeddings directly from the output tokens of the image encoder.
For open-vocabulary object classification, the embeddings computed by the class prediction head are compared to text encoder embeddings of object descriptions.
This architecture achieves strong open-vocabulary object detection performance~\cite{minderer2022simple,minderer2024scaling} and does not suffer from the training instabilities observed in some encoder-decoder detection models~\cite{detr,minderer2022simple, zhu2021deformable, yao2021efficient, song2022vidt, zhang2022dino}.

\subsection{Extension to Relationship Prediction}

In the architecture described above, each image encoder token represents an object proposal that captures per-object information. 
Importantly, the encoder consists of self-attention layers which, by design, model all pairwise interactions between these tokens. 
We therefore hypothesize that information about object relationships can be learned directly in the image encoder.
To extract this information in the form of \triplet{} triplets, we introduce two MLPs that transform vision encoder output tokens into \subject{} embeddings $\mathbf{s}_i$ and \object{} embeddings $\mathbf{o}_j$ (\Cref{fig:architecture}). 
Using different MLPs for \subject{} and \object{} is crucial to break symmetry in subsequent processing (\Cref{app:hparams}).

An embedding representing the relationship between two object proposals (encoder tokens) can then be obtained simply by element-wise addition of their \subject{} and \object{} embeddings. 
Obtaining relationship embeddings for all $N^2$ \tuple{} pairs, where $N$ is the number of object proposals, would be computationally infeasible.
We therefore introduce a \emph{Relationship Attention} layer that performs hard attention to select the \tuple{} pairs most likely to form a relationship. This layer computes an attention-like score $p_{ij} = \mathbf{s}_i\mathbf{o}_j^T$, where $\mathbf{s}_i$ and $\mathbf{o}_j$ are the embedding vectors of \subject{}~$i$ and \object{}~$j$, and $p_{ij}$ represents the likelihood that a relationship between \subject{}~$i$ and \object{}~$j$ exists.
By computing this score for all \tuple{} pairs, we obtain an $N{\times}N$ matrix, from which we select the top $k$ pairs.\footnote{This selection reduces the number of relationships that need to be processed, e.g., from $N^2 = 3600^2$ to $k=16386$ (99.9\% reduction) for our ViT-L/14 model.}
For these $k$ pairs, we compute relationship embeddings $\mathbf{r}_{ij} = \textrm{LayerNorm}(\mathbf{s}_i + \mathbf{o}_j)$. 
To classify the relationship of a pair, its relationship embedding is compared to text embeddings of relationship predicates.
We use the embeddings where \subject{} and \object{} are identical ($\mathbf{r}_{i=j}$) to represent object instances, and predict object classes from them. Boxes are predicted from the corresponding image encoder output tokens (see \Cref{fig:architecture}).

The Relationship Attention layer therefore identifies object pairs to focus on for relationship classification, by computing a hard attention with \subject{} embeddings as queries and \object{} embeddings as keys. 
Since the hard selection is not differentiable, gradients from the relationship prediction cannot be used directly to train this layer. Instead, the relationship score is self-supervised to predict the maximum \predicate{} class probability that will later be predicted for the relationship at the classification stage (\Cref{sec:training}).

\subsection{Training}\label{sec:training}

The image and text encoders of our model are initialized from a vision-language model that was contrastively pretrained on a large number of image-text pairs~\cite{radford2021learning}. After adding the Relationship Attention as well as the object bounding box and class prediction heads, the model is jointly trained on a mixture of object and relationship detection datasets in a single training stage (\Cref{sec:experimental-setup}), using the losses described below.

\textbf{Bipartite Matching.}
Bipartite matching between object predictions ($\mathbf{r}_{i=j}$) and ground-truth annotations is performed based on a cost consisting of the object classification loss and the box prediction losses as in~\cite{detr}.
This matching between objects also establishes a matching of predicted to ground-truth relationship predicates, since a relationship is uniquely identified by the \subject{} and \object{} indices $i$ and $j$.
Unmatched predictions are trained to predict low scores for all classes and incur no box prediction loss. 

\textbf{Box Prediction Losses.} For bounding box regression, we use the L1 and generalized intersection-over-union (gIoU) losses described in~\cite{detr}.

\textbf{Object and Predicate Classification Loss.}
For both object category and predicate classification, we use a sigmoid cross-entropy loss as in~\cite{minderer2022simple}. This loss is computed between the ground truth classes and logits obtained by computing the inner product of the relationship embeddings $\mathbf{r}_{ij}$ selected by the Relationship Attention layer with the text embeddings of the class names.
For embeddings corresponding to individual objects ($\mathbf{r}_{i = j}$), the class name is the object category or description.
For embeddings corresponding to relationships ($\mathbf{r}_{i \ne j}$), we use the predicate text as class name.
This differs from prior work~\cite{univrd, yuan2023rlipv2, yuan2022rlip}, which uses the full \triplet{} triplet description.
Our approach has the advantage of disentangling object category and predicate names, which allows for more efficient inference since object category and predicate texts are embedded separately.

\textbf{Relationship Score Loss.}
To select only the most promising embeddings for further processing, the Relationship Attention layer predicts a score $p_{ij}$ that represents the likelihood that the corresponding \texttt{<subject-object>} pair forms a relationship (if $i \ne j$), or that the corresponding object exists in the image (if $i=j$).
This score is trained with a sigmoid cross-entropy loss against targets provided by the model itself, namely the maximum probability predicted for any class for the corresponding $\mathbf{r}_{ij}$ embedding~\cite{minderer2024scaling}.
In this way, the relationship score is trained to predict the class probability of a potential embedding $\mathbf{r}_{ij}$ \emph{before} actually computing and classifying the embedding.
Note that this loss can only be computed for those objects and relationships that ultimately get selected for further processing.
We found this to provide sufficient supervision.

The final loss is an equally weighted sum of four components: (1) classification loss, (2) L1 box loss, (3) gIoU box loss, (4) relationship score loss.

\newcommand{\rowA}{\rowcolor{gray!30}}
\newcommand{\rowB}{\rowcolor{gray!15}}
\newcommand{\celG}[1]{\textcolor{gray!90}{#1}}
\newcommand{\numD}[1]{\textcolor{red}{$_{\mathbf{{\downarrow #1}}}$}}
\newcommand{\numU}[1]{\textcolor{blue}{$_{\mathbf{{\uparrow #1}}}$}}

\section{Experiments}
\label{sec:experiments}

\subsection{Experimental Setup}
\label{sec:experimental-setup}

\vspace{4mm}
\subsubsection{Datasets.}
We use the following datasets for training or evaluation:

\textit{Visual Genome}~\cite{krishna2017visual} is the largest VRD dataset, labeled with 2.3M triplet relationships across 108K images. 
However, as Visual Genome includes noisy annotations, the community commonly evaluates VRD on a cleaned version of the dataset~\cite{zellers2018neural} where the label space is reduced to the 150 most frequent object classes and the 50 most frequent predicate classes. This dataset is commonly referred to as Visual Genome 150 (VG150). 

\textit{GQA}~\cite{hudson2019gqa} uses the same image corpus as Visual Genome, but is more diversely labeled, with 1703 object classes (1704 including a background class) and 310 predicates. Spatial relationships like ``to the left of'' are automatically labeled based on the 2-D spatial arrangement of bounding boxes. Similar to VQ150, GQA200 is a reduced and cleaned version of GQA with 200 object classes and 100 predicate classes~\cite{dong2022stacked}.

\textit{HICO-DET}~\cite{chao2018} is a dataset which focuses on a more specialized subset of visual relationships, specifically interactions between humans and objects.
HICO-DET contains ~50k images and is exhaustively annotated for 600 defined human-object interaction (HOIs). 

\textit{Objects365}~\cite{shao2019objects365} (O365) is a large scale object-detection dataset with 365 object categories across two million images.

\textit{Open X-Embodiment}~\cite{open_x_embodiment_rt_x_2023} (OXE) dataset is targeted towards learning vision language action policies for robotics. OXE lacks bounding box annotations which precludes quantitative evaluation, but it represents both a visually different setting compared to the standard VRD benchmarks and a promising application area, so we use it for qualitative evaluation.

\subsubsection{Training.} Unless otherwise noted, we train models on a mixture of VG, VG150, GQA200, HICO, and O365 in the proportions show in \Cref{tab:dedup}. The B/32 model is trained for 200'000 steps at batch size 256 (further details in \Cref{app:hparams}). Since the datasets we use share similar image sources, there is some overlap between the official training and evaluation splits. To ensure that no evaluation images are used for training, we rigorously filter all training datasets to remove images present in any of the test splits for all datasets we evaluate on, i.e. VG150, GQA, GQA200, and HICO. We use an image similarity filter that also identifies non-exact matches~\cite{kolesnikov2020big}. \Cref{tab:dedup} shows the number of images removed from each training split. Note that not all prior work performs similar deduplication.

\begin{table}[t]
\scriptsize
\begin{center}
\begin{tabularx}{0.95\linewidth}{l|*{6}{C}}
\toprule
& VG150 & VG  & HICO & GQA200 & GQA  & Objects365 \\ \midrule
\% in training mixture\hspace{1em} & $12.5\%$ & $12.5\%$ & $12.5\%$ & $12.5\%$ & $0\%$ & $50\%$\\
\# removed\hspace{1em} & $6587$ & $17727$ & $729$ & $15361$ & $18801$ & $15306$ \\ \bottomrule
\end{tabularx}
\end{center}
\vspace{-1mm}
\caption{\textbf{Training data mixture.} \emph{\# removed} indicates the number of images that were removed from the official training split due to overlap with data we evaluate on.}
\label{tab:dedup}
\vspace{-8mm}
\end{table}

\subsubsection{Evaluation Procedure and Metrics.}
Prior work presents a diverse range of evaluation metrics, often tailored to specific datasets. For non-exhaustively labelled datasets, like Visual Genome and GQA, using precision-based metrics is inherently inconclusive. Thus, the community has adapted a \textit{Recall}@$K$ metric for these datasets, where $K$ denotes a fixed budget of \triplet{} triplets on which the recall is computed. However, this metric is biased towards the more frequent \predicate{} classes in the dataset~\cite{tang2019learning, chen2019knowledge}. Therefore, recent approaches use \textit{mean Recall}@$K$, which calculates \textit{Recall}@$K$ separately for each \predicate{} class in the test data and then averages the results.

For exhaustively labeled datasets like HICO, it is feasible to use precision-based metrics such as \textit{mean Average Precision} (mAP), a well-established metric in object detection. For HICO, the mean is taken over the 600 possible HOI triplets. Besides the overall metric, we also report results separately for rare ($<10$ occurrences) and non-rare ($\ge10$ occurrences) HOIs as mAPr and mAPn.

Further, evaluation can be ``graph-constrained'', where only a single prediction can be made per detected object pair, or ``graph-unconstrained'', where any number of predictions can be made per pair. 
Existing works evaluating on Visual Genome and GQA datasets partially lack a clear distinction between graph-constrained and unconstrained evaluation. For our models, we report graph-constrained results in the main paper, which reflects the dominant approach used in other studies. Results without graph-constraint are reported in \Cref{app:graph_unconstrained}. 
For the HICO dataset, we follow the established evaluation procedure~\cite{chao2018} that inherently represents a graph-unconstrained evaluation.

Multiple implementations of the aforementioned metrics exist, often showing differences in results. To compare fairly and directly against a wide range of prior approaches we identified the PyTorch evaluation procedure of~\cite{tang2020unbiased} as the most prevalent routine for recall-based metrics and replicated their procedure in JAX with numerical accuracy. Similarly, for HICO, we numerically reproduce the original Matlab evaluation procedure outlined in \cite{chao2018} in JAX.

\subsubsection{Exhaustive Relationship Evaluation.} Previous methods~\cite{univrd, yuan2022rlip, yuan2023rlipv2} entangle objects and relationships in a single representation. Consequently, these methods score their embeddings against the full \triplet{} triplet. While this is feasible for datasets with a low number of possible triplets, like HICO with 600 defined relationships, it quickly becomes computationally intractable for datasets with a larger vocabulary of objects and predicates (e.g. VG, GQA). 
Prior works~\cite{univrd} therefore evaluate only on triplets which are known to be present in the test split, which may inflate metrics. In contrast, our method disentangles objects and predicates (\Cref{sec:model}). This allows for an exhaustive evaluation that considers all object and predicate combinations and enables applications where knowledge of possible relationships is unavailable.

\subsection{Relationship Detection Performance}
\label{sec:vg-results}

\begin{table}[t]
\scriptsize
\parbox{1.0\linewidth}{
\centering
\begin{tabular}{llccc}
\toprule
Model & Backbone & mR@$50$ & mR@$100$\\
\midrule
RelTR~\cite{cong2023reltr} & ResNet-50 & $6.8$ & $10.8$ \\
Transformer + CFA~\cite{li2023compositional} & ResNeXt-101 & $12.3$ & $14.6$\\
VCTree~\cite{tang2019learning} + IETrans + Rwt~\cite{zhang2022fine} & ResNeXt-101 & $12.0$ & $14.9$\\
Motif~\cite{zellers2018neural} + IETrans + Rwt~\cite{zhang2022fine} & ResNeXt-101 & $15.5$  & $18.0$\\
GPS-Net~\cite{lin2020gps} + IETrans + Rwt~\cite{zhang2022fine} & ResNeXt-101 & $16.2$ & $18.8$\\
SG-Transformer~\cite{yu2020cogtree} + IETrans + Rwt~\cite{zhang2022fine} & ResNeXt-101 & $15.0$ & $18.1$\\
Sgtr~\cite{li2022sgtr} & ResNet-101 & $15.8$ & $20.1$ \\
DT2-ACBS~\cite{desai2021learning} & ResNet-101 & $22.0$ & $24.4$\\
\midrule
UniVRD & CLIP: ViT-B/32 & $9.6$ & $12.1$\\
UniVRD & CLIP: ViT-B/16 & $10.9$ & $13.2$\\
UniVRD & CLIP: ViT-L/14 & $12.6$ & $14.5$\\
\midrule
SG-ViT & CLIP: ViT-B/32 & $15.0$ & $18.1$ \\
SG-ViT & CLIP: ViT-B/16 & $15.7$ & $19.3$ \\
SG-ViT & CLIP: ViT-L/14 & $17.8$ & $21.8$\\
SG-ViT + simple predicate rebalancing & CLIP: ViT-L/14 & $\bf{22.3}$ & $\bf{26.1}$\\
\bottomrule
\end{tabular}
\vspace{2mm}
\caption{\textbf{Performance on the Visual Genome 150 test set.} We use graph-constrained evaluation (one predicate prediction per \texttt{<subject-object>} pair). For the last row (\emph{``+ simple text rebalancing''}), the model was briefly fine-tuned on VG150 data with rebalanced predicate frequencies (see \Cref{sec:vg-results,app:hparams}).}
\label{tbl:vg_sota}}
\vspace{-8mm}
\end{table}

We use Visual Genome as our main benchmark for relationship detection.
\Cref{tbl:vg_sota} shows results for our models and prior work.
Prior studies often achieve reported performance through the combination of a base model with a specialized approach to overcome the skewed data distribution (indicated as \textit{base-model} + \textit{specialization} in \Cref{tbl:vg_sota}).
Our models provide strong performance even without special treatment of the skewed predicate distribution. 
After brief fine-tuning on data in which the frequency of predicate annotations was rebalanced with simple rejection sampling (see \Cref{app:hparams}), our method improves on the prior best method DT2-ACBS~\cite{desai2021learning} by 1.7 points mR@100 (26.1 vs. 24.4). This shows that our already-strong method may be further improved by combining it with more advanced rebalancing methods.

We also note the large difference in performance to UniVRD~\cite{univrd}, which builds on the same detection architecture as our method, but adds a Transformer decoder-based relationship module.
To disentangle whether the improvements over UniVRD are due to differences in architecture or training data, we also trained our B/32 model on the UniVRD data mixture, and still observe a large improvement (17.4 VG mR@100 for our method vs. 12.1 for UniVRD when trained on the same data; \Cref{tbl:dataset_ablation}).
This suggests that our encoder-only architecture is better suited for VRD than prior decoder-based architectures~\cite{univrd, cong2023reltr, kim2021hotr, li2022sgtr}.

\begin{table}[t]
\scriptsize
\parbox[b]{0.47\linewidth}{
\centering
\begin{tabular}{lcc}
\toprule
Model & mR@$50$ & mR@$100$\\
\midrule
LLM4SGG~\cite{kim2023llm4sgg} & $5.3$ & $6.5$\\
Neural Motif w/ GCL~\cite{dong2022stacked} & $16.8$ & $18.8$\\
VCTree w/ GCL~\cite{dong2022stacked} & $15.6$ & $17.8$\\
SG-Transformer w/ CFA~\cite{li2023compositional} & $13.4$ & $15.3$\\
SHA w/ GCL~\cite{dong2022stacked} & $17.8$ & $20.1$\\
\midrule
SG-ViT (CLIP: ViT-B/32) &  $15.7$ &  $18.6$ \\
SG-ViT (CLIP: ViT-B/16) & $17.7$ & $20.5$ \\
SG-ViT (CLIP: ViT-L/14) & $\mathbf{19.3}$ & $\mathbf{22.9}$ \\
\bottomrule
\end{tabular}
\vspace{1mm}
\caption{\textbf{Performance on the GQA200 test set.} Graph-constrained evaluation.}
\label{tbl:gqa200_sota}
}
\hfill
\parbox[b]{0.47\linewidth}{
\centering
\begin{tabular}{lcc}
\toprule
Model & mR@$50$ & mR@$100$\\
\midrule
\rowA \multicolumn{3}{l}{\textit{Scene Graph Classification}}\\
IMP~\cite{xu2017scene, koner2020relation, suhail2021energy} & $0.5$ & $0.7$\\
Neural Motif~\cite{zellers2018neural, koner2020relation, suhail2021energy} & $0.8$ & $1.2$\\
Unbiased TDE~\cite{tang2020unbiased, koner2020relation} & $-$ & $0.7$ \\
RTN~\cite{koner2020relation} & $-$ & $1.4$ \\
MP~\cite{knyazev2020graphdensity} & $-$ & $2.8$\\
Transformer w/ EBM~\cite{suhail2021energy} & $1.3$ & $1.8$\\
\midrule
\rowA \multicolumn{3}{l}{\textit{Scene Graph Generation}}\\
SG-ViT (CLIP: ViT-B/32) & $6.2$ & $7.4$ \\
SG-ViT (CLIP: ViT-B/16) & $6.4$ & $7.4$ \\
SG-ViT (CLIP: ViT-L/14) & $\mathbf{8.0}$ & $\mathbf{9.6}$ \\
\bottomrule
\end{tabular}
\vspace{1mm}
\caption{\textbf{Performance on the GQA test set.} Graph-constrained evaluation.}
\label{tbl:gqa_sota}
}
\vspace{-8mm}
\end{table}

\subsection{Scaling to Large Vocabularies}
\label{sec:open-vocab-results}

Improving long-tail performance has been critical to the VRD field since the natural distribution of relationship triplets is highly skewed~\cite{LI2024127052}. A substantial body of literature is devoted to this goal, often proposing complex loss debiasing or data augmentation approaches~\cite{LI2024127052,yu2020cogtree,zhang2022fine,desai2021learning}.
It is therefore of interest whether our method works well in this regime without special treatment of rare classes.
To evaluate the performance of our method on large object and predicate vocabularies with a long tail of rare classes, we use the Graph Question Answering (GQA) dataset~\cite{hudson2019gqa}. This dataset builds on VG and expands the number of classes and annotations. 
A simplified version of GQA with 200 object and 100 predicate classes, called GQA200~\cite{dong2022stacked}, is part of our training mixture.
Our method surpasses prior results on the GQA200 test split (\Cref{tbl:gqa200_sota}), despite lacking the specialized treatment of the data distribution that some prior methods use.

The full GQA dataset has an even larger vocabulary with 1703 object and 311 predicate classes.
To our knowledge, no prior methods evaluate scene graph \emph{generation} on the full GQA dataset. Some prior methods report results on scene graph \emph{classification}, where ground-truth boxes are provided. As shown in \Cref{tbl:gqa_sota}, our model achieves higher performance on the much harder task of scene graph \emph{generation} than prior methods on \emph{classification}. We further assess zero-shot generalization to unseen classes in \Cref{app:gqa-unseen}.

We believe that two factors contribute to the strong performance in the open vocabulary regime:
First, the simple design of the method does not impede transfer of semantic knowledge from the pre-trained VLM backbone because it adds only lightweight heads.
In particular, we use no decoder, which could be difficult to train and could lead to ``forgetting'' of pretrained representations.
Instead, most of the relationship modeling happens in the backbone, which has had the benefit of large-scale vision-language pretraining.
Second, the open-vocabulary design allows end-to-end training on a mix of datasets, which was not possible for all prior methods.

\subsection{Human-Object Interaction}
\label{sec:hoi-results}
\afterpage{\begin{table}[t]
\begin{center}
\scriptsize
\begin{tabular}{llccc}
\toprule
Model & Backbone & mAP & mAPr & mAPn\\
\midrule
HOTR~\cite{kim2021hotr} & ResNet50 & $25.10$ & $20.33$ & $25.86$ \\
RLIP~\cite{yuan2022rlip} & ResNet50 & $32.84$ & $26.85$ & $34.63$\\
RLIPv2~\cite{yuan2023rlipv2} & ResNet50 & $35.38$ & $29.61$ & $37.10$\\
RLIPv2~\cite{yuan2023rlipv2} & Swin-L & $\mathbf{45.09}$ & $\mathbf{43.23}$ & $\mathbf{45.64}$\\
UniVRD~\cite{Zhao_2023_ICCV} & CLIP: ViT-B/32 & $29.98$ & $22.94$ & $32.02$\\
UniVRD~\cite{Zhao_2023_ICCV} & CLIP: ViT-B/16 & $29.98$ & $22.94$ & $32.02$\\
UniVRD~\cite{Zhao_2023_ICCV} & CLIP: ViT-L/14 & $37.41$& $28.90$ & $39.95$\\
\hline
SG-ViT & CLIP: ViT-B/32 & $28.86$ & $23.72$ & $30.39$ \\
SG-ViT & CLIP: ViT-B/16 & $31.98$ & $26.83$ & $33.52$ \\
SG-ViT & CLIP: ViT-L/14 & $38.11$ & $33.71$ & $39.42$\\
\bottomrule
\end{tabular}
\end{center}
\caption{\textbf{Performance on the HICO test set.} Our method is on par with the most comparable prior work, UniVRD, for overall mAP. However, our method performs better on ``rare'' classes (mAPr), indicating better generalization to rarely seen concepts. For reference, we also show RLIPv2, the state-of-the-art method on HICO. RLIPv2 proposes a self-training approach that is orthogonal and compatible to our method.}
\label{tbl:hico}
\end{table}
}

We also evaluate our model on the specialized task of human-object interaction, using the HICO benchmark~\cite{chao2018}.
The performance of our model is on par with the most comparable prior method~\cite{univrd}, but does not show similarly large improvements as for VG or GQA200 (\Cref{tbl:hico}).
HICO has a much narrower and more specialized vocabulary than VG or GQA (HICO evaluates on 600 pre-specified triplets, whereas VG150 has $1.1\times10^6$ possible triplet combinations).
Performance on this specialized task may benefit less from transfer of pretrained representations and may instead be limited by the amount of task-specific training data.
This is supported by the fact that the recent state-of-the-art method for HICO pretrains on large amounts of person-focused pseudo-labels~\cite{yuan2023rlipv2}.

\subsection{Ablations}
\label{sec:ablations}

\begin{wrapfigure}{rt}{0.3\textwidth} 
    \vspace{-7.6mm}
    \centering
    \includegraphics[width=\linewidth]{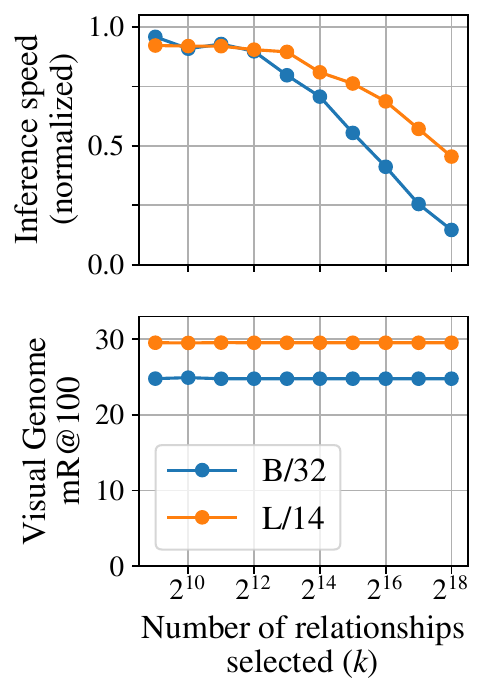}
    \vspace{-4mm}
    \caption{Model speed and VRD accuracy by number of selected relationships $k$. Speed is relative to a non-VRD object detector~\cite{minderer2022simple}.}
    \label{fig:inference-speed}
    \vspace{-7.6mm}
\end{wrapfigure}

\vspace{3mm}
\subsubsection{Inference Speed and Number of Predicted Relationships.}

In the following, we ablate training datasets and model components to study the interplay of data and architecture on the model performance.

The primary hyperparameter of our method is the number of relationships $k$ that are selected by the Relationship Attention layer from all possible \texttt{<subject-object>} combinations for further processing.
\Cref{fig:inference-speed} shows how VRD performance and inference speed depend on $k$. 

For training, we use $k=2^{14}$, which results in 71\% (B/32) to 81\% (L/14) of the speed of a pure object detection model~\cite{minderer2022simple} of the same size.
Without top-$k$ selection, the model would be several times slower (right end of top plot in \Cref{fig:inference-speed}).
Since the vast majority of \texttt{<subject-object>} pairs do not form relationships, VRD performance is essentially unaffected by top-$k$ selection at inference.
For inference, we can therefore use $k=2^{11}$ at no loss in accuracy to achieve 90\% of the speed of the object detector (for B/32 and L/14). 
At this $k$, the B/32 model achieves 52.8 FPS at batch size 1 on an NVIDIA V100 GPU and is therefore suitable for real-time applications.

\subsubsection{Dataset Mixture.} To assess how training data and architectural improvements contribute to the performance of our model compared to prior work, we train the models on a range of dataset mixtures (\Cref{tbl:dataset_ablation}). The full mixture, which we use for all experiments unless otherwise noted, includes VG, VG150, GQA200, HICO and O365.
To allow direct comparison to UniVRD~\cite{univrd}, we also train on their mixture (VG150, HICO, O365, COCO) and find that our model still performs better on VG150 by a large margin (18.1 mR@100 vs 12.1 for UniVRD for B/32 models). Since UniVRD uses the same image and text encoding backbones as our model, this result suggests that our encoder-only model with Relationship Attention is an advance over decoder-based relationship architectures.
We further find that including both pure object detection (O365) and large-vocabulary VRD data (VG) benefit all metrics.

\begin{table}[t]
\scriptsize
\parbox{1.0\linewidth}{
\centering
\begin{tabular}{l|cccccc|c|c|c}
\toprule
 & \multicolumn{6}{c|}{Mixture} & VG150 & GQA200 & HICO\\
& VG & VG150 & GQA200 & HICO & O365 & COCO & mR@$100$ & mR@$100$ & mAP \\
\midrule
Our mixture &  $12.5\%$ & $12.5\%$ & $12.5\%$ & $12.5\%$ & $50\%$ & $-$ & $18.1$ & $18.6$ & $28.9$ \\
\midrule
UniVRD\cite{univrd} mix & $-$ & $10\%$ & $-$ & $20\%$ & $50\%$ & $20\%$ & $17.4\hspace{0.3em}\downarrow\hspace{0.5em} 4\%$ & $\hspace{0.5em}5.1\hspace{0.3em}\downarrow 73\%$ & $ 17.4\hspace{0.3em}\downarrow\hspace{0.5em} 8\%$  \\
ours w/o O365 &  $25\%$ & $25\%$ & $25\%$ & $25\%$ & $-$ & $-$ & $15.3\hspace{0.3em}\downarrow 15\%$ & $15.6\hspace{0.3em}\downarrow 16\%$ & $23.8\hspace{0.3em}\downarrow 18\%$  \\
ours w/o VG &  $-$ & $25\%$ & $12.5\%$ & $12.5\%$ & $50\%$ & $-$ & $ 14.2\hspace{0.3em}\downarrow 22\%$ & $15.3\hspace{0.3em}\downarrow 18\%$ & $26.8\hspace{0.3em}\downarrow\hspace{0.5em} 7\%$  \\
\bottomrule
\end{tabular}
\vspace{3mm}
\caption{\textbf{Dataset Ablation.\label{tbl:dataset_ablation}} Performance of SG-ViT (B/32) trained on different data mixtures. VG150 and GQA200 evaluation is graph-constrained, HICO is unconstrained.}
}
\end{table}

\subsubsection{Object Detection Performance.} A necessary step towards good relationship detection performance is accurate object detection. In our architecture, object detection is treated as a special case of relationship detection in which the \subject{} is identical to the \object{}.  We validate this design by training the model only on detection datasets without relationship annotations, using the same dataset mixture as the OWL-ViT~\cite{minderer2022simple} model (Objects365 and Visual Genome, \cite{minderer2022simple}). We find that our model achieves a similar object detection performance as OWL-ViT (21.9\% vs. 22.1\% mAP and 18.3\% vs. 18.9\% mAPr on LVIS~\cite{gupta2019lvis}), indicating that the Relationship Attention design does not interfere with object detection.

\begin{figure}[ht!]
\setlength\tabcolsep{3pt}
\renewcommand{\arraystretch}{0.}
\centering
\begin{tabular}{@{} p{16mm} M{0.24\linewidth} M{0.24\linewidth} M{0.24\linewidth} @{}}
\textbf{Relation} & \textbf{GT} & \textbf{SG-ViT (B/32)} & \textbf{SG-ViT (L/14)} \\ \addlinespace
\begin{subfigure}{\linewidth} \caption{\\\raggedright building near track}\label{subfig:building_near_track} \end{subfigure} 
  & \includegraphics[width=\hsize]{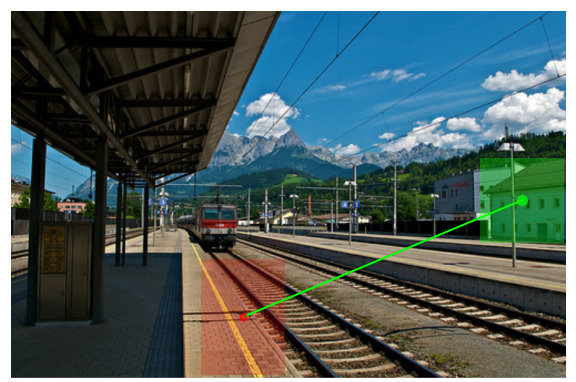}  \vspace{-4mm}
  & \includegraphics[width=\hsize]{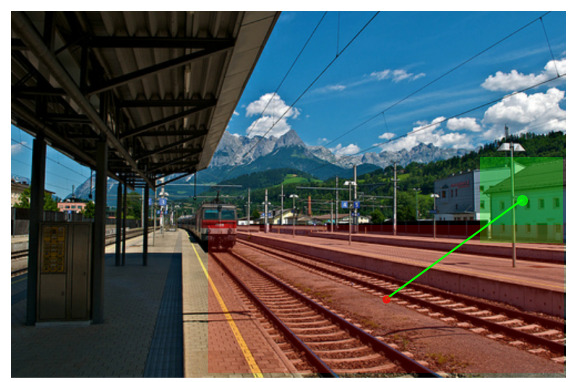}  \vspace{-4mm}
  & \includegraphics[width=\hsize]{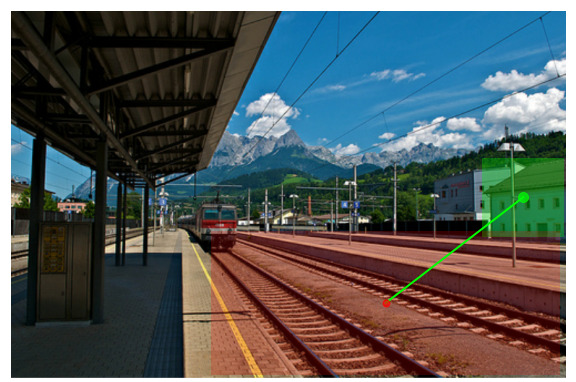}  \vspace{-4mm}\\ 
\begin{subfigure}{\linewidth} \caption{\\\raggedright light on bus}\label{subfig:light_on_bus} \end{subfigure} 
  & \includegraphics[width=\hsize]{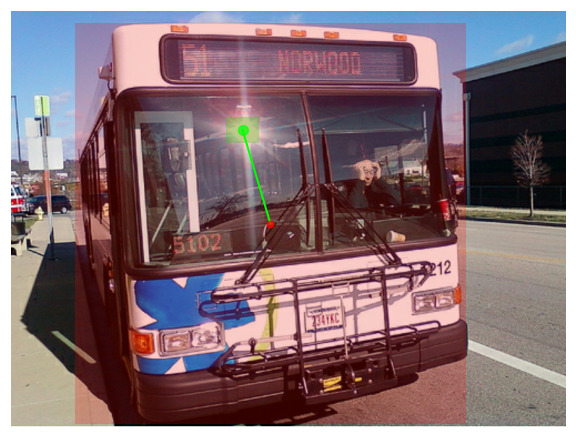}  \vspace{-4mm}
  & \includegraphics[width=\hsize]{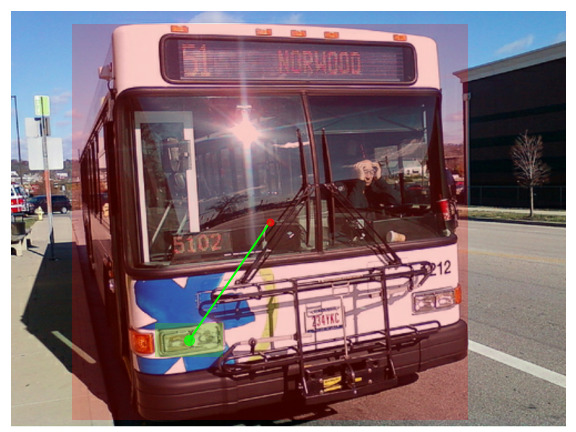}   \vspace{-4mm}
  & \includegraphics[width=\hsize]{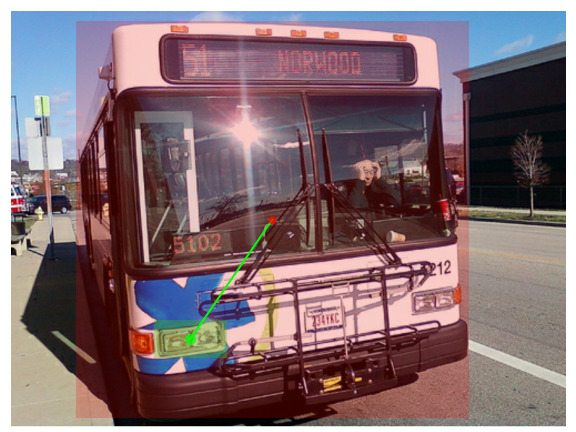}  \vspace{-4mm}\\
\begin{subfigure}{\linewidth} \caption{\\\raggedright snow on hill}\label{subfig:snow_on_hill} \end{subfigure} 
  & \includegraphics[height=20mm]{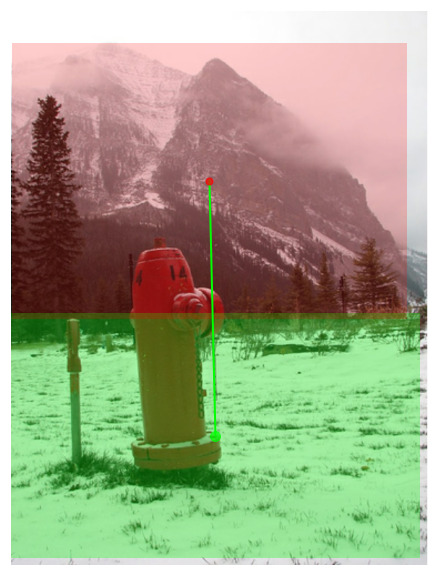} 
  & \includegraphics[height=20mm]{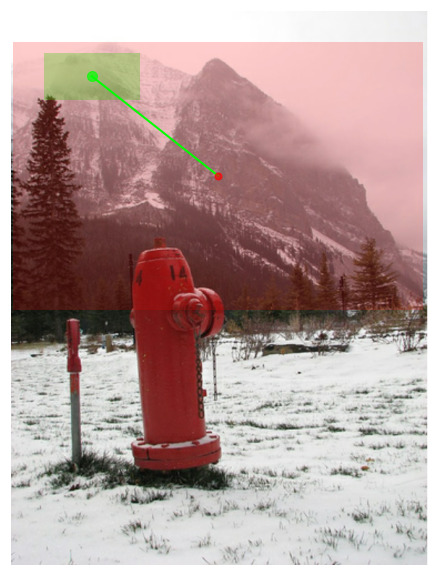} 
  & \includegraphics[height=20mm]{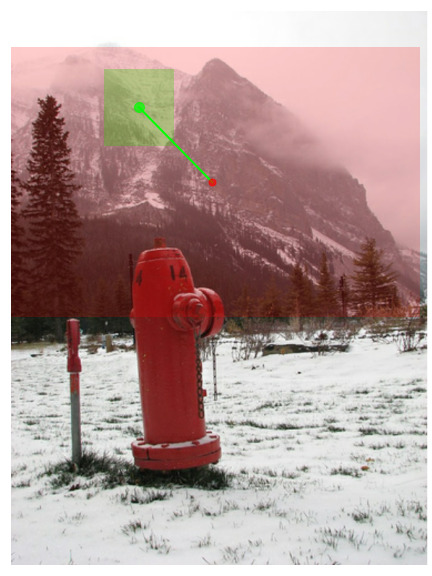} \\
\begin{subfigure}{\linewidth} \caption{\\\raggedright bottle in door}\label{subfig:bottle_in_door} \end{subfigure} 
  & \includegraphics[height=20mm]{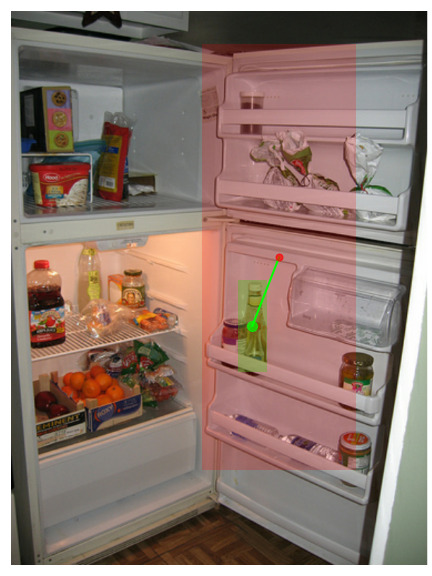} 
  & \includegraphics[height=20mm]{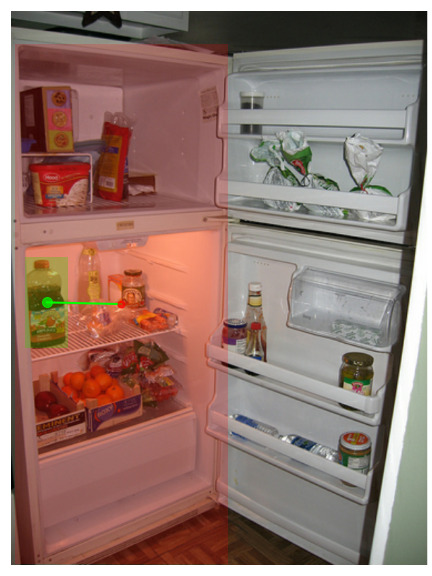} 
  & \includegraphics[height=20mm]{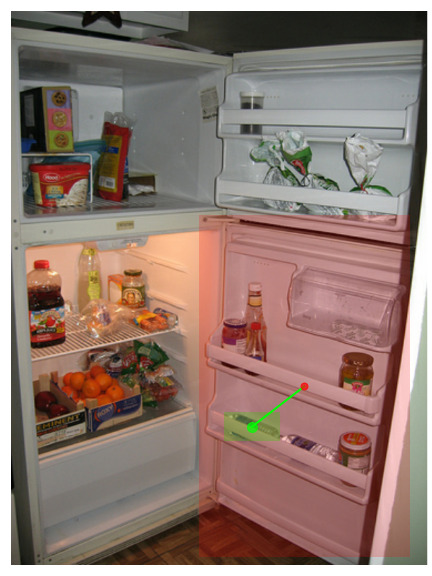} \\
\end{tabular}
\caption{Qualitative examples of difficult edge-cases from VG150 test split \cite{krishna2017visual, zellers2018neural}. From left to right: Ground Truth, SG-ViT (B/32), SG-ViT (L/14). In all cases the \subject~is \textcolor{green}{lime} and the \object~is \textcolor{red}{red}.}
\label{fig:sg-vit-qualitative-results}
\end{figure}

\begin{figure}[ht]
    \centering
    \begin{subfigure}[b]{\textwidth}
        \includegraphics[width=0.29\textwidth]{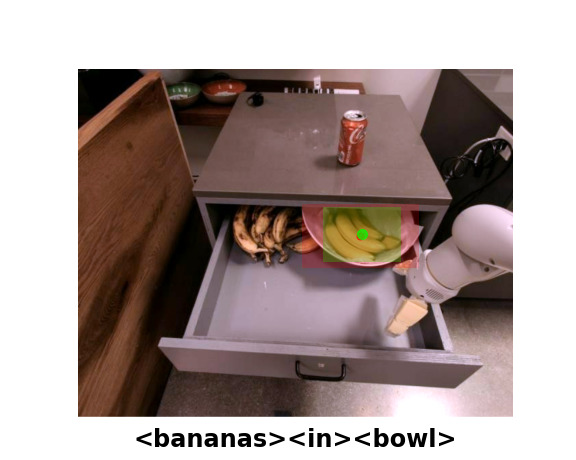}
        \hspace{-8mm}
        \includegraphics[width=0.29\textwidth]{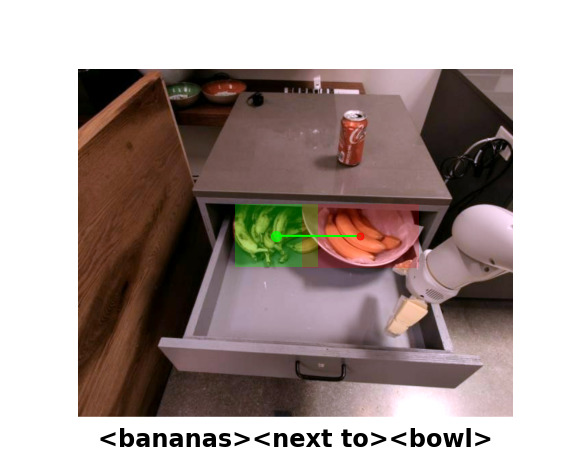} 
        \hspace{-8mm}
        \includegraphics[width=0.29\textwidth]{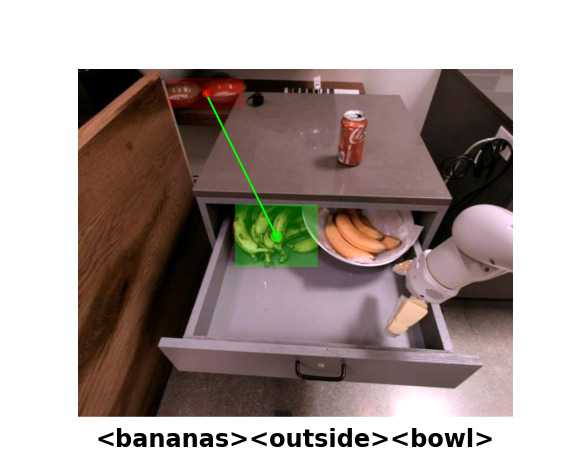} 
        \hspace{-8mm}
        \includegraphics[width=0.29\textwidth]{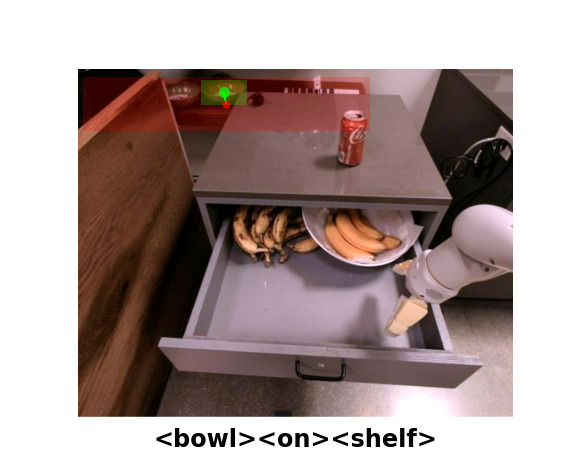} 
    \vspace{-7.5mm}
    \end{subfigure}
    \begin{subfigure}[b]{\textwidth}
        \includegraphics[width=0.29\textwidth]{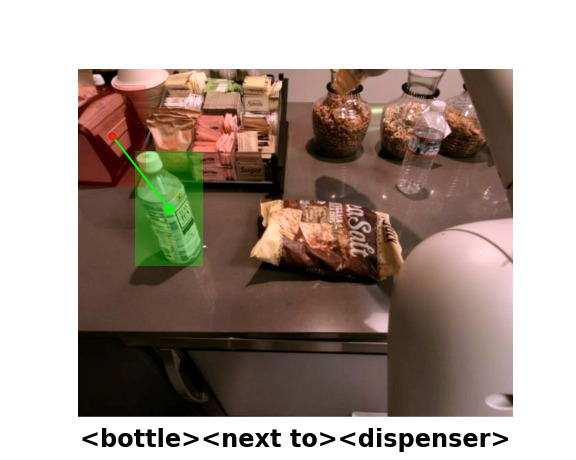}
        \hspace{-8mm}
        \includegraphics[width=0.29\textwidth]{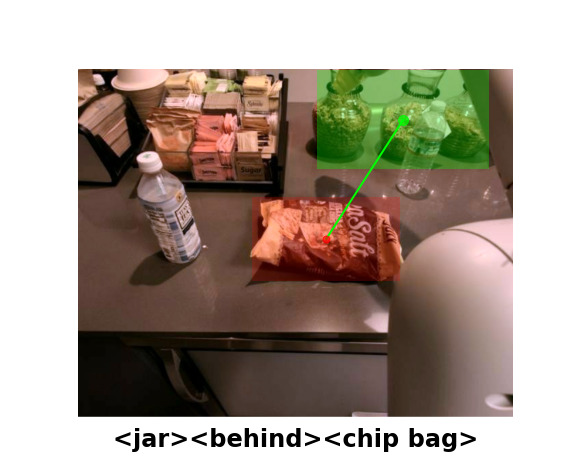}
        \hspace{-8mm}
        \includegraphics[width=0.29\textwidth]{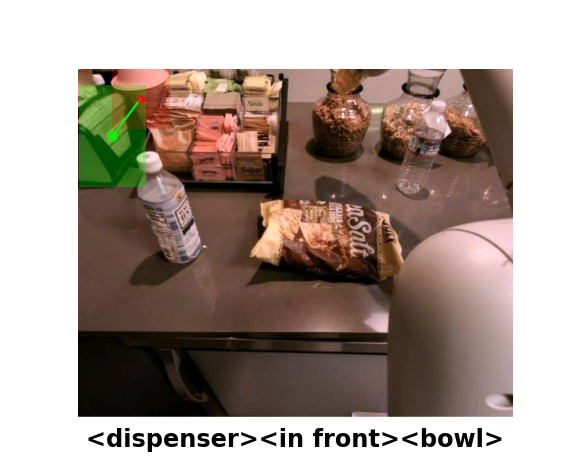}  
        \hspace{-8mm}
        \includegraphics[width=0.29\textwidth]{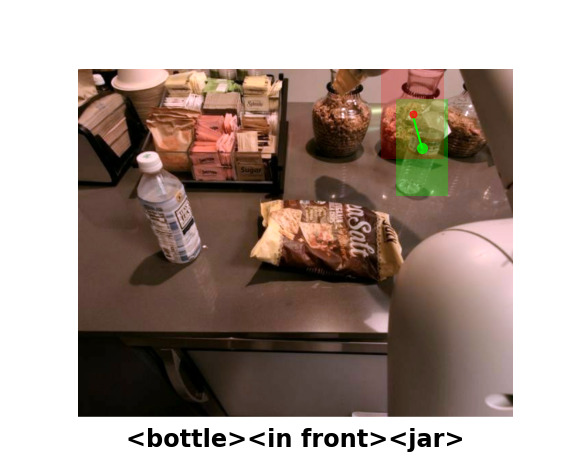} 
    \vspace{-4.5mm}
    \end{subfigure}
    
    \caption{Qualitative examples showing SG-ViT (L/14) on out-of-distribution data from the OXE dataset~\cite{open_x_embodiment_rt_x_2023}. In all cases the \subject~is \textcolor{green}{lime} and the \object~is \textcolor{red}{red}.
    Note how the model correctly disambiguates several instances of the same class (e.g. ``bananas'' and ``bottle'') depending on their relationships.
    }
    \label{fig:sg-vit-qualitative-results-robotics}
\end{figure}

\subsection{Qualitative Results}
\label{sec:qualitative-results}

In \Cref{fig:sg-vit-qualitative-results}, we visualize qualitative examples to highlight difficult edge cases.
Even where the model output differs from the ground truth, it is often plausibly correct, e.g. where either the extent of an object differs with the ground truth or there are multiple appropriate, but unannotated, relations. For example, both models select the headlight on the bus in \Cref{subfig:light_on_bus} for ``light on bus'' as opposed to the sun reflected off the windshield. In \Cref{subfig:snow_on_hill} both models focus more on the \predicate{~``on''} in ``snow on hill'' than the distinction between a mountain and a hill. \Cref{subfig:bottle_in_door} shows a challenging scene where the B/32 model selects the bottle inside the open fridge while the L/14 correctly selects a water bottle in the door for ``bottle in door'' while also capturing the extent of the door instance.

Furthermore, to qualitatively assess the utility of this work beyond standard relation benchmark datasets, we also visualize some example relations applied to data from the robotics domain, namely the OXE dataset \cite{open_x_embodiment_rt_x_2023}. \Cref{fig:sg-vit-qualitative-results-robotics} shows a real-life scene where multiple instances of the same semantic object categories are present within the scene. The model is able to match the \predicate{} to the instances in the desired configuration. One subjective failure-case is that the ``banana outside bowl'' text query had its highest scoring triplet involving a bowl on a shelf in the background rather than the adjacent bowl. Another characteristic is that the model occasionally places boxes over groups with multiple instances of the same semantic class when a singular word is used for either the \object{} or \subject{}. This can be attributed to the similarity between word and their plural in the CLIP embeddings and that many human annotations from VG also confuse singular and plural (see \Cref{app:examples}).

\section{Limitations}
While our model performs strongly on large-vocabulary VRD, its performance on specialized human-object interaction detection is only on par with prior models. We discuss this limitation in \Cref{sec:hoi-results}.
Further error modes are explored qualitatively in \Cref{sec:qualitative-results} and \Cref{app:examples}.
A challenge for open-vocabulary relationship detection models as a whole is zero-shot generalization to unseen objects and predicates. While our model improves over prior approaches (\Cref{tbl:gqa200_sota,tbl:gqa_sota}), there is still a large gap when it comes to entirely unseen classes (\Cref{app:gqa-unseen}). Future research should focus on closing this gap.

\section{Conclusion}
We present an architecture for open-vocabulary visual relationship detection.
By combining an encoder-only design with a novel Relationship Attention layer for efficient selection of high-confidence relationships, our architecture leverages VLM pretraining and multi-dataset VRD training. It achieves strong performance on standard and large-vocabulary VRD benchmarks while maintaining pure object detection performance and adding little extra computational cost. Due to its simplicity and strong performance, we believe that our method will be a useful basis for further research on visual relationship detection.

\bibliographystyle{splncs04}
\bibliography{references}

\clearpage
\appendix
\section{Appendix}

\subsection{Model Details and Hyperparameters}
\label{app:hparams}

Unless otherwise noted, we follow the public implementation\footnote{\scriptsize{\url{https://github.com/google-research/scenic/tree/main/scenic/projects/owl_vit}}} of~\cite{minderer2022simple}. Below, we provide additional details.

\subsubsection{Relationship Attention Architecture.}

\begin{wrapfigure}{rt}{0.3\linewidth} 
    \vspace{-8mm}
    \centering
    \includegraphics[width=1\linewidth]{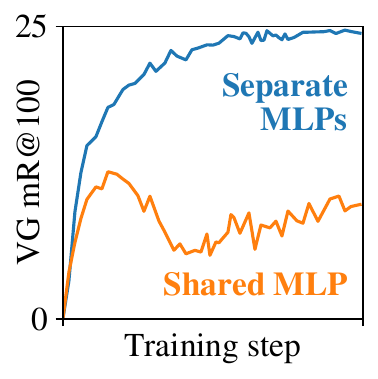}
    \vspace{-5mm}
    \caption{Computing \subject{} and \object{} embeddings with separate MLPs is necessary for good performance.}
    \label{app:fig:breaking-symmetry}
    \vspace{-5mm}
\end{wrapfigure}

To obtain query, key, and value embeddings from the image embeddings for the Relationship Attention layer, we use 3-layer MLPs with no change in feature dimensionality, GeLU hidden activations~\cite{hendrycks2016gaussian}, and a skip connection from the input to the output embedding. LayerNorm~\cite{ba2016layer} is applied to the final output in the MLPs. To obtain the relationship embedding, \subject{} and \object{} embeddings are summed, normalized by LayerNorm, and processed by another 2-layer MLP (not shown in \Cref{fig:architecture}). We found model performance to be robust to the details of the Relationship Attention layer, e.g. the hyperparameters of the MLPs, and it may be possible to simplify the design further.
However, as noted in the main paper, \subject{} and \object{} embeddings must be computed with different projections to model the asymmetry between \subject{} and \object{} in the relationship.
If the same projection were used to compute a single embedding to represent both \subject{} and \object{}, the model could not distinguish between e.g.\ \texttt{"person riding horse"} and \texttt{"horse riding person"}. \Cref{app:fig:breaking-symmetry} confirms experimentally that sharing the MLP for subjects and objects performs poorly. With a shared MLP, the model struggles to learn and reaches a low final score.

\subsubsection{Relationship Selection.}
In the Relationship Attention layer, we perform two rounds of top-$k$ selection as depicted in \Cref{fig:architecture}: First, we select the top 512 object instances, using the diagonal entries of the relationship score matrix as ``objectness'' scores. This reduces the size of the relationship score matrix from $N \times N$ (where $N$ is the number of image encoder output tokens, i.e. object proposals) to $512 \times 512$. From this matrix, we then select $k$ relationships, where $k = 2^{14} = 16384$ unless otherwise noted. Additionally, we always compute embeddings for the 512 self-relationships along the diagonal (which represent object instances), since these embeddings will be necessary to classify the object categories of the \subject{} and \object{} boxes. Model performance is remarkably robust to the value of $k$ both during training and inference. We did not observe a significant reduction in performance for $k$ as low as 1024 either just during inference (\Cref{sec:ablations}) or during training and inference.

\subsubsection{Data Augmentation.}
For data preprocessing and augmentation, we follow~\cite{minderer2022simple} with some exceptions to account for the differences between general object detection and relationship detection data: For object detection datasets, we apply random left/right flip and random crop augmentation, up to $3 \times 3$ mosaics, and random negative labels. For relationship detection datasets, we replace the random crop with random resizing between $0.5 \times$ and $1.0 \times$ of the original size, since cropping may cause label inaccuracies if one member of a relationship is cropped off. For GQA200, we also remove random left/right flipping since the dataset contains spatial relationship annotations in the form of ``\subject{} to the left of \object{}''. We do not use random prompt templates such as \texttt{"a photo of a \{\}"} or prompt ensembling for any datasets.

\subsubsection{Data Rebalancing.}
For most of our experiments, we use the training data as-is without special treatment of the skewed object or predicate class distributions. Only for the model on the last line in \Cref{tbl:vg_sota}, we perform simple rebalancing as follows, to show the potential of combining our method with orthogonal approaches focusing on data rebalancing: We first count the number of occurrences of each predicate in the training set to obtain its frequency. Then, during training, we randomly drop relationship annotations with a probability equal to its frequency (capped at 0.95 for the most frequent predicates).

\subsubsection{Training Details.}
The B/32 and B/16 models are trained on images of size $768 \times 768$ at batch size 256 for 200'000 steps with the Adam optimizer~\cite{kingma2014adam} and a cosine learning rate schedule with a maximal learning rate of $5 \times 10^{-5}$ and a 1000-step linear warmup. As in~\cite{minderer2022simple}, the text encoder is trained with a learning rate of $2 \times 10^{-6}$ instead. For the L/14 model, the image size is $840 \times 840$, batch size is 128, and the maximal learning rate is $2 \times 10^{-5}$.

\subsubsection{Speed Benchmarking.}
For the speed benchmarking in \Cref{fig:inference-speed}, we assume a scenario in which a stream of images (e.g. a video feed) needs to be processed with a fixed set of 1000 text queries (i.e. 1000 object and predicate classes). We therefore report the time needed to process a new image, given pre-computed text query embeddings. We measure the time from calling the model with a single image (batch size 1) until the predictions are ready, using an NVIDIA V100 GPU. We measure 30 trials and report the median result.

\subsection{Additional Experimental Results}
\vspace{3mm}

\subsubsection{Zero-shot GQA.}\label{app:gqa-unseen}
To assess zero-shot generalization to unseen classes, we report the performance on the least-frequent 1503 object and 211 predicate classes in GQA, i.e. those \emph{not} included in GQA200 and therefore unseen during training (\Cref{tbl:gqa-unseen}). Given the large vocabulary and the difficulty of zero-shot predicate classification, we evaluate the model without graph-constraint, allowing four predicate predictions per \texttt{<subject-object>} pair. 
Although the performance of our model in this scenario is nontrivial, it is significantly lower than on seen classes (\Cref{tbl:gqa_sota}).
We therefore suggest using the least-frequent GQA annotations in this manner as a challenging benchmark for future work on zero-shot VRD.

\subsubsection{Recall@K.}
We provide results using the Recall@K metric (i.e. pooling all classes before recall computation) in \Cref{tbl:recall}. Note that this metric weighs classes by their frequency and is therefore not suitable for assessing long-tail performance~\cite{tang2019learning, chen2019knowledge}.

\subsubsection{Performance without Graph-Constraint.} \label{app:graph_unconstrained}
All results in the main paper for VG150 and GQA are computed with graph constraint. \Cref{tbl:unconstrained} provides these results \emph{without} graph constraint.

\vspace{5mm}
\begin{table}
\centering
\begin{tabular}{lcc}
\toprule
Model & mR@$50$ & mR@$100$\\
\midrule
\rowA \multicolumn{3}{l}{\textit{Scene Graph Generation (unseen classes)}}\\
SG-ViT (CLIP: ViT-B/32) & $1.5$ & $2.2$ \\
SG-ViT (CLIP: ViT-B/16) & $1.9$ & $2.3$ \\
SG-ViT (CLIP: ViT-L/14) & $\mathbf{2.2}$ & $\mathbf{2.8}$ \\
\bottomrule
\end{tabular}
\vspace{1mm}
\caption{\textbf{Performance on the GQA test set (unseen classes only).} Evaluated \emph{without} graph-constraint.}
\label{tbl:gqa-unseen}
\end{table}

\begin{table}
\scriptsize
\centering
\begin{tabular}{l|ccc|ccc}
\toprule
 & \multicolumn{3}{c|}{Visual Genome 150} & \multicolumn{3}{c}{GQA200} \\
 & R@$20$ & R@$50$ & R@$100$ & R@$20$ & R@$50$ & R@$100$  \\
\midrule
SG-ViT (CLIP: ViT-B/32) & $19.8$ & $28.1$ & $34.5$ & $16.4$ & $22.9$ & $27.9$ \\
SG-ViT (CLIP: ViT-B/16) & $20.2$ & $28.8$ & $35.4$ & $16.6$ & $23.4$ & $28.9$ \\
SG-ViT (CLIP: ViT-L/14) & $21.8$ & $31.1$ & $38.3$ & $18.6$ & $26.6$ & $32.6$ \\
\bottomrule
\end{tabular}
\vspace{3mm}
\caption{\textbf{Evaluation on Recall metrics.} Evaluated without graph-constraint.}
\label{tbl:recall}
\end{table}

\begin{table}
\scriptsize
\centering
\begin{tabular}{l|cc|cc}
\toprule
 & \multicolumn{2}{c|}{Visual Genome 150} & \multicolumn{2}{c}{GQA200} \\
 & mR@$50$ & mR@$100$ & mR@$50$ & mR@$100$\\
\midrule
SG-ViT (CLIP: ViT-B/32) & $20.5$ & $24.8$ & $21.9$ & $26.1$ \\
SG-ViT (CLIP: ViT-B/16) & $21.4$ & $26.6$ & $23.2$ & $27.4$ \\
SG-ViT (CLIP: ViT-L/14) & $23.9$ & $29.5$ & $26.7$ & $32.8$ \\
\bottomrule
\end{tabular}
\vspace{3mm}
\caption{\textbf{Evaluation without graph-constraints.}}
\label{tbl:unconstrained}
\end{table}

\clearpage
\subsection{Additional Qualitative Examples}
\label{app:examples}
\begin{figure}
\setlength\tabcolsep{3pt}
\centering
\begin{tabular}{@{} p{16mm} M{0.24\linewidth} M{0.24\linewidth} M{0.24\linewidth} @{}}
\textbf{Relation} & \textbf{GT} & \textbf{SG-ViT (B/32)} & \textbf{SG-ViT (L/14)} \\ 
\begin{subfigure}{\linewidth} \caption{\\\raggedright bottle on sink}\label{subfig:bottle on sink} \end{subfigure} 
  & \includegraphics[width=\hsize]{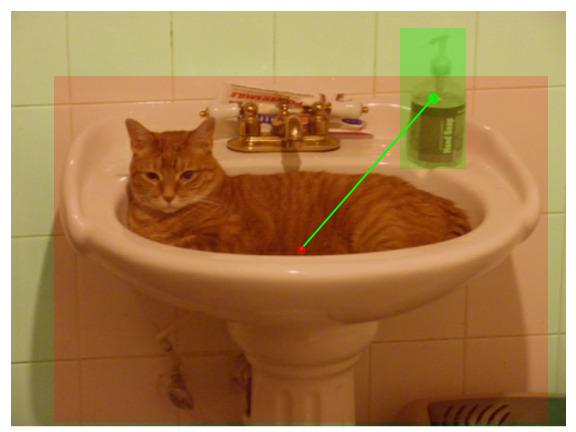} \vspace{-5mm}
  & \includegraphics[width=\hsize]{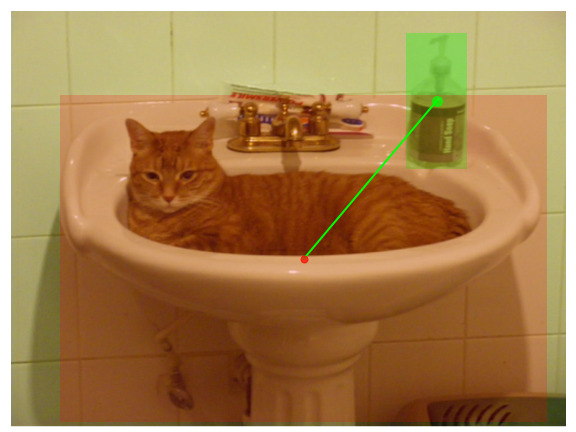} \vspace{-5mm}
  & \includegraphics[width=\hsize]{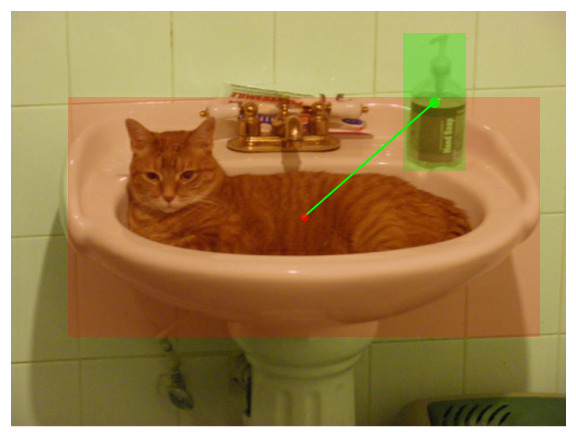} \vspace{-5mm}\\
\begin{subfigure}{\linewidth} \caption{\\\raggedright number on post}\label{subfig:number_on_post} \end{subfigure} 
  & \includegraphics[width=\hsize]{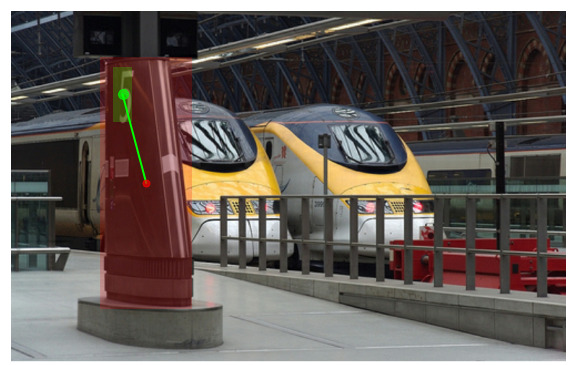}  \vspace{-5mm}
  & \includegraphics[width=\hsize]{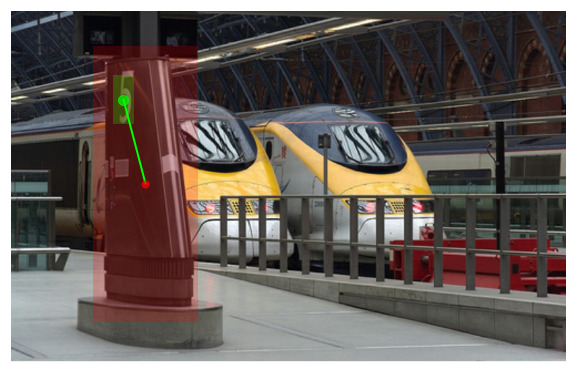}   \vspace{-5mm}
  & \includegraphics[width=\hsize]{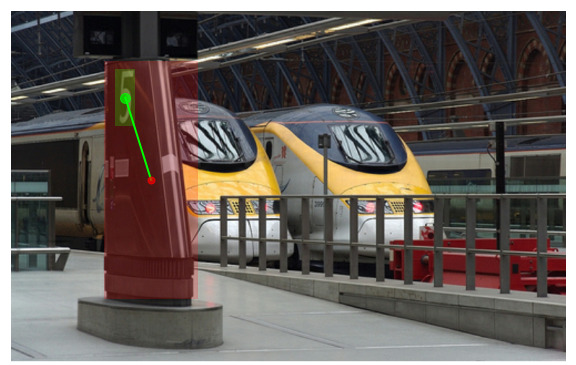}  \vspace{-5mm}\\
\begin{subfigure}{\linewidth} \caption{\\\raggedright light of bike}\label{subfig:light_of_bike} \end{subfigure} 
  & \includegraphics[width=\hsize]{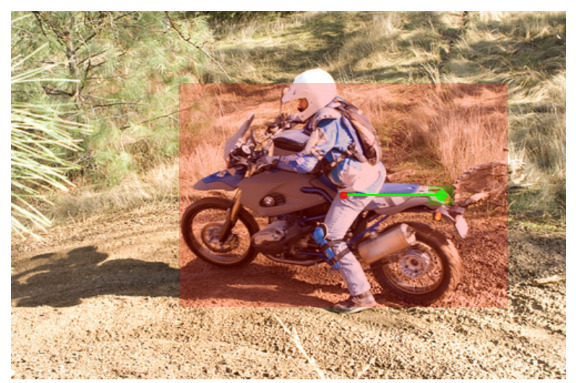} \vspace{-5mm}
  & \includegraphics[width=\hsize]{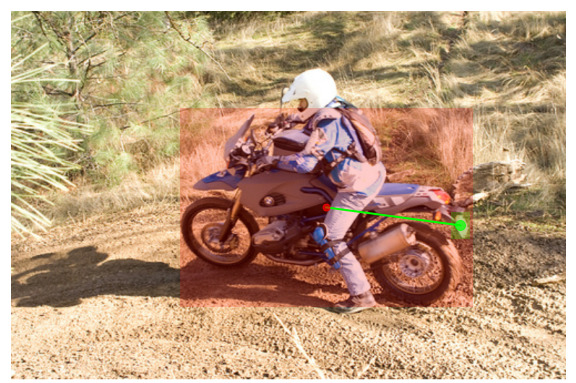} \vspace{-5mm}
  & \includegraphics[width=\hsize]{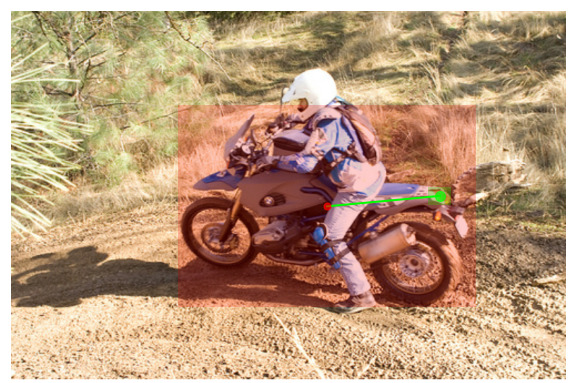} \vspace{-5mm}\\
\begin{subfigure}{\linewidth} \caption{\\\raggedright fruit on table}\label{subfig:fruit_on_table} \end{subfigure} 
  & \includegraphics[width=\hsize]{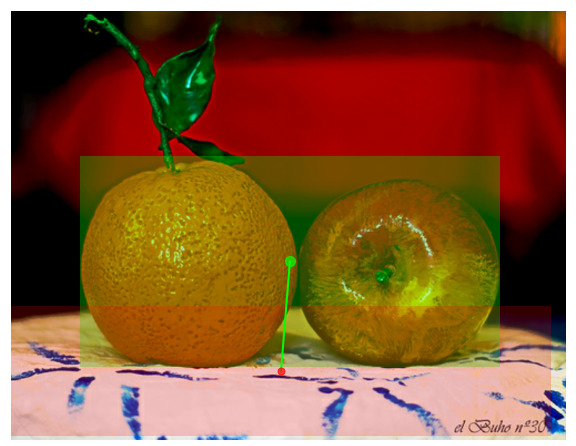} \vspace{-5mm}
  & \includegraphics[width=\hsize]{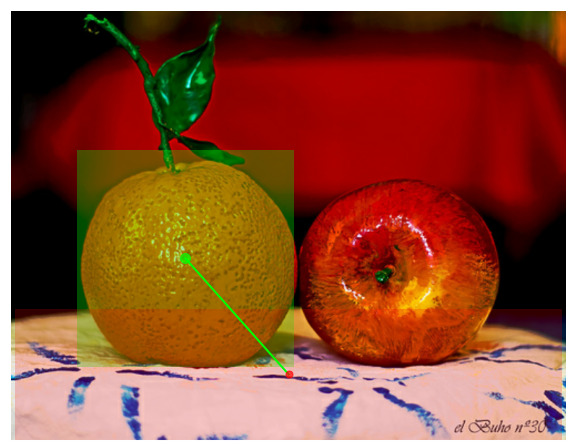} \vspace{-5mm}
  & \includegraphics[width=\hsize]{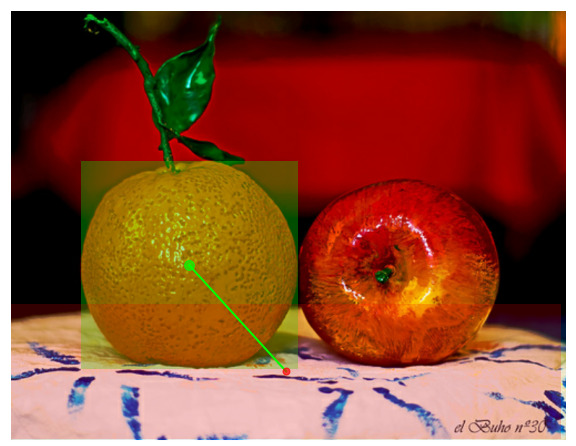} \vspace{-5mm}\\
\begin{subfigure}{\linewidth} \caption{\\\raggedright plane with wing}\label{subfig:plane_with_wing} \end{subfigure} 
  & \includegraphics[width=\hsize]{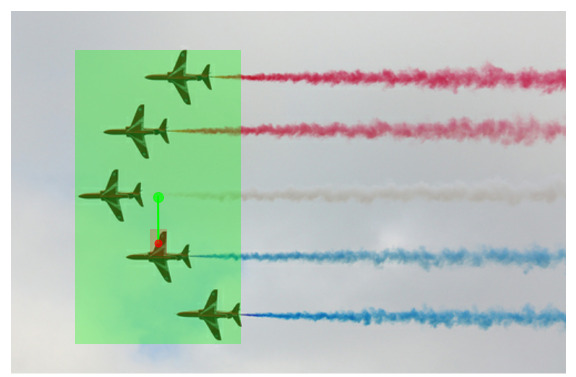}  \vspace{-5mm}
  & \includegraphics[width=\hsize]{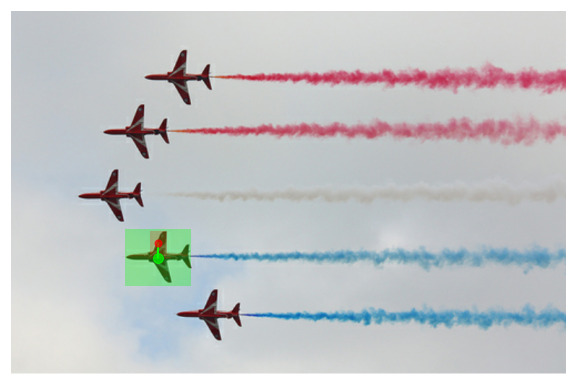}  \vspace{-5mm}
  & \includegraphics[width=\hsize]{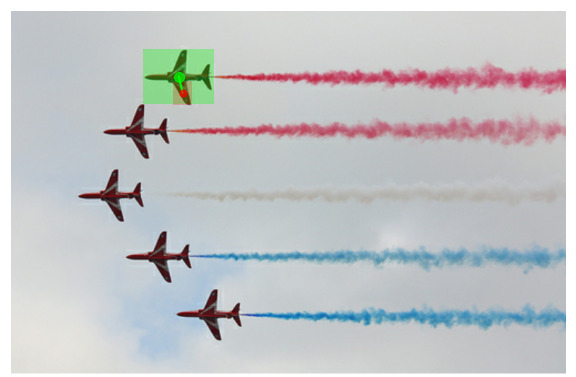}  \vspace{-5mm}\\
\begin{subfigure}{\linewidth} \caption{\\\raggedright food on plate}\label{subfig:food_on_plate} \end{subfigure} 
  & \includegraphics[width=\hsize]{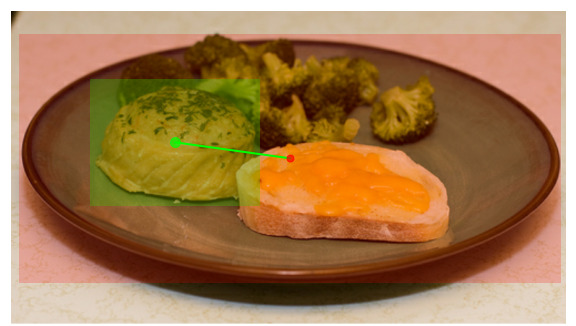} \vspace{-5mm}
  & \includegraphics[width=\hsize]{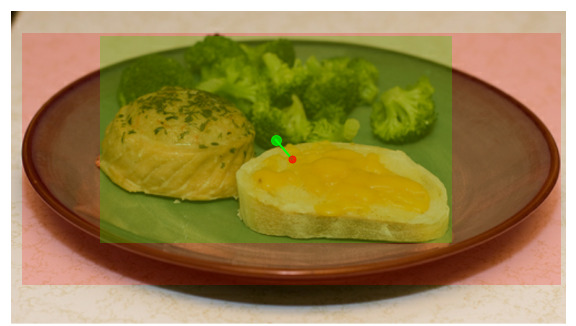} \vspace{-5mm}
  & \includegraphics[width=\hsize]{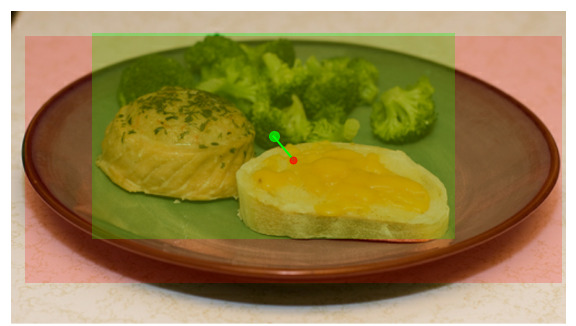} \vspace{-5mm}\\
\begin{subfigure}{\linewidth} \caption{\\\raggedright snow on mountain}\label{subfig:snow_on_mountain} \end{subfigure} 
  & \includegraphics[width=\hsize]{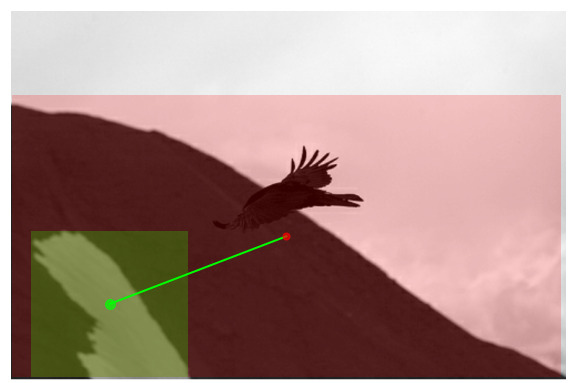}  \vspace{-5mm}
  & \includegraphics[width=\hsize]{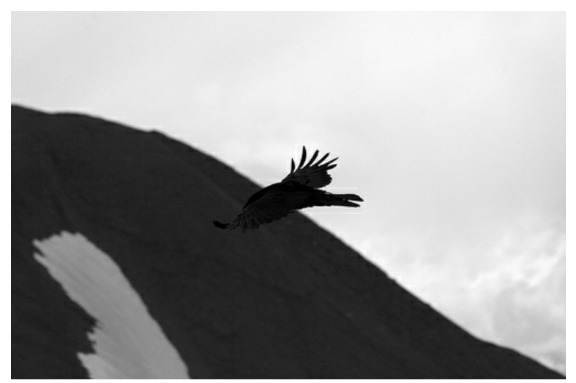}  \vspace{-5mm}
  & \includegraphics[width=\hsize]{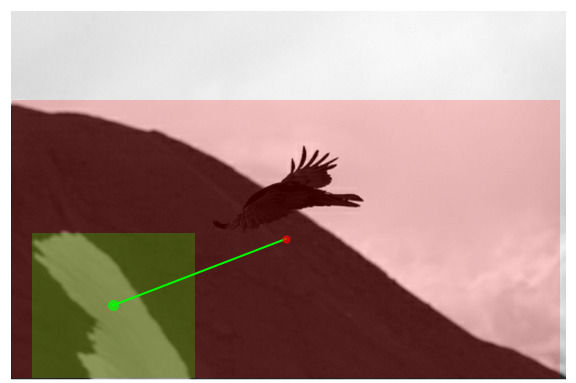}  \vspace{-5mm} \\
\begin{subfigure}{\linewidth} \caption{\\\raggedright elephant near giraffe}\label{subfig:elephant_near_giraffe} \end{subfigure} 
  & \includegraphics[width=\hsize]{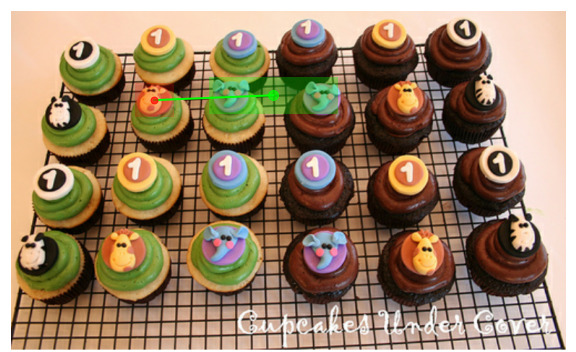}  \vspace{-5mm}
  & \includegraphics[width=\hsize]{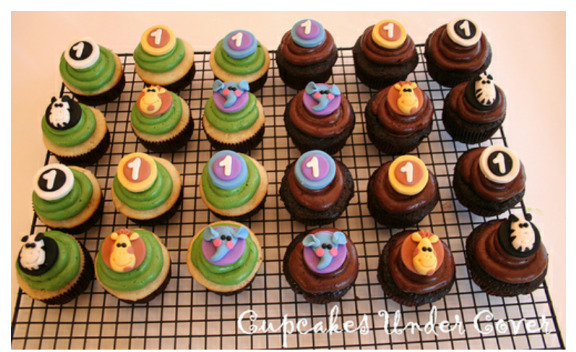}  \vspace{-5mm}
  & \includegraphics[width=\hsize]{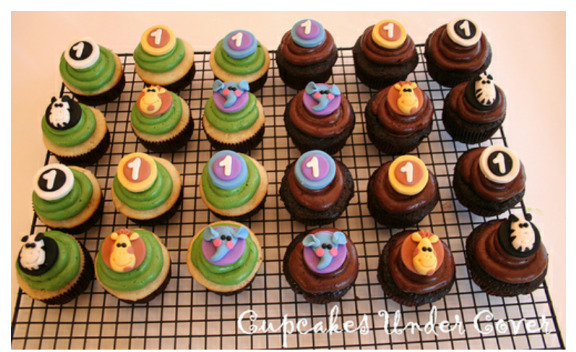}  \vspace{-5mm}\\ 
\end{tabular}
\vspace{-2mm}
\caption{Additional qualitative examples from VG150 test split \cite{krishna2017visual, zellers2018neural} illustrating box outputs of the SG-ViT B/32 and L/14 models for the given text. Rows (a-b) show true positives with only minor differences in the object bounding boxes. Row (c) shows a minor error in subject from the B/32 model. Rows (d-g) illustrate the challenge of selecting singular or plural entities while rows (g-h) show examples where the relationship score is very low (i.e. no output if triplet score < 0.001). In all cases the \subject~is \textcolor{green}{lime} and the \object~is \textcolor{red}{red}.}
\label{fig:sg-vit-qualitative-results-additional}
\end{figure}

\Cref{fig:sg-vit-qualitative-results-additional} shows additional qualitative examples of relationships predicted by SG-ViT on VG150 and illustrates some error modes. The bounding boxes for each node of the relationship edge accurately captures the extent of the object instances while in each case aligning the grammatical \subject{} and \object{} in the correct direction, denoted by the shaded boxes \textcolor{green}{lime} and \textcolor{red}{red} respectively. The only false positive in this set of images is in \Cref{subfig:light_of_bike} where the B/32 model scores the relationship ``light of bike'' with the subject box selecting the bright licence plate of the motorcycle instead of the brake light.

A more frequent error mode is the confusion of singular instances and groups of instances. In some cases this can be put down to ambiguity in language. For example in \Cref{subfig:fruit_on_table} the subject text ``fruit'' could be referring to super-set category that includes both oranges and apples and also shares the same word for both singular and plural. In other cases we see that this confusion also appears in the human annotations, e.g. in \Cref{subfig:plane_with_wing} and \Cref{subfig:food_on_plate}. Such inconsistency is likely common in the training set, which may impact the model's ability to distinguish singular and plural.

False negatives are another error mode, in which the model predicts no boxes, or assigns very low confidence to predictions. Two such examples are shown in \Cref{subfig:snow_on_mountain} and \Cref{subfig:elephant_near_giraffe} where the relationship descriptions are ``snow on mountain'' and ``elephant near giraffe'' respectively. On these examples, the model predicts no relationships with a score above 0.001, which we use as a threshold for visualization for all examples shown here. In the first case, the snow is only recognized by L/14 model and for the latter case neither model recognizes the decorations on the cup-cakes as either an elephant or giraffe.

\subsection{Additional Graph Visualization Examples}
\begin{figure}
    \centering
    \begin{subfigure}{\textwidth}
        \centering
        \includegraphics[height=1.4in]{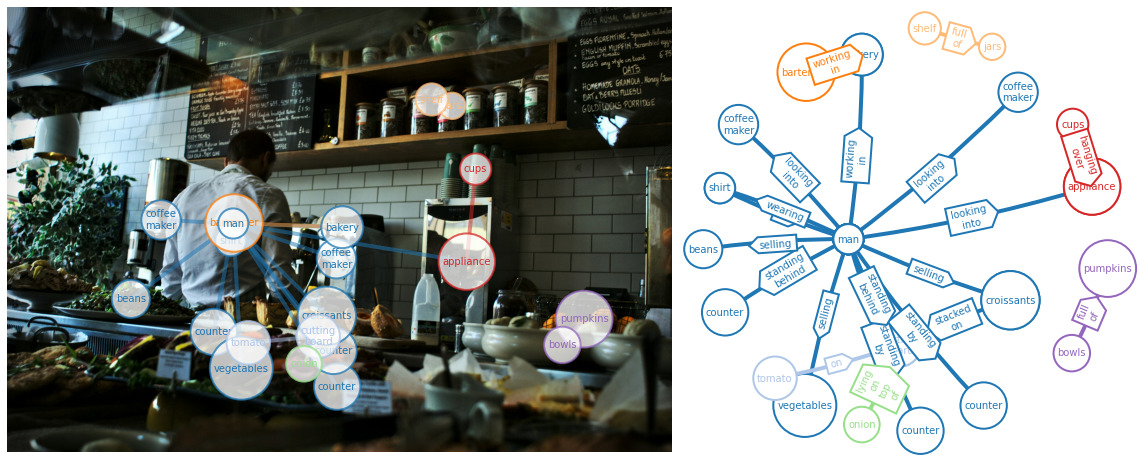}
        \caption{Photo by Joanna Boj on \href{https://unsplash.com/photos/man-standing-beside-espresso-maker-MhOoD_h90ks}{Unsplash}.}
    \end{subfigure}
    \begin{subfigure}{\textwidth}
        \centering
        \includegraphics[height=1.4in]{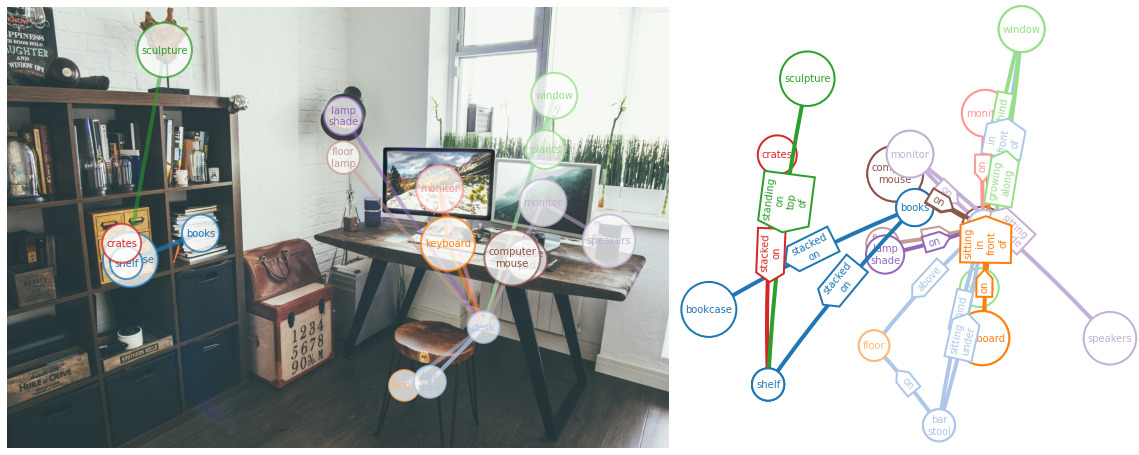}
        \caption{Photo by Vadim Sherbakov on \href{https://unsplash.com/photos/two-flat-screen-monitor-turned-on-near-organizer-rack-inside-the-room-RcdV8rnXSeE}{Unsplash}.}
    \end{subfigure}
    \begin{subfigure}{\textwidth}
        \centering
        \includegraphics[height=1.4in]{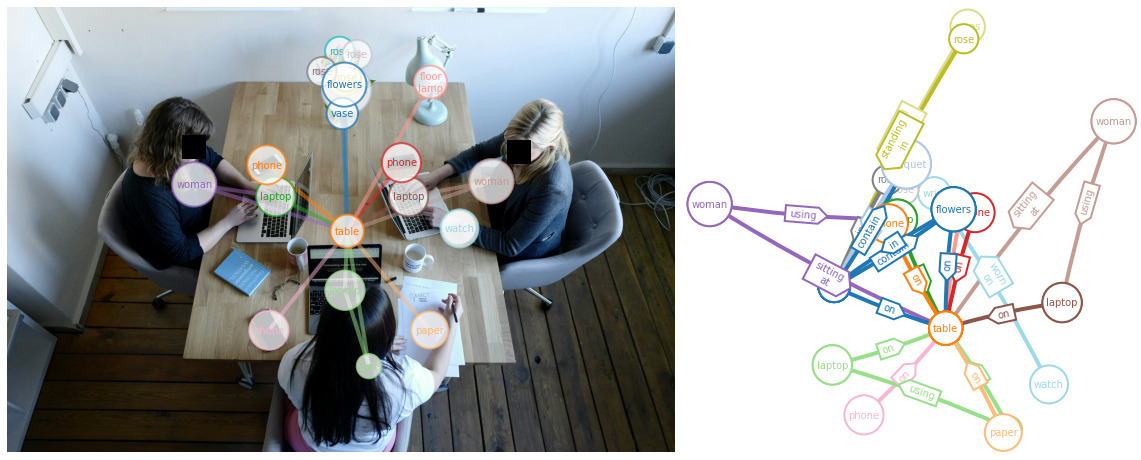}
        \caption{Photo by CoWomen on \href{https://unsplash.com/photos/three-women-sitting-around-table-using-laptops-7Zy2KV76Mts}{Unsplash}.}
    \end{subfigure}
    \begin{subfigure}{\textwidth}
        \centering
        \includegraphics[height=1.4in]{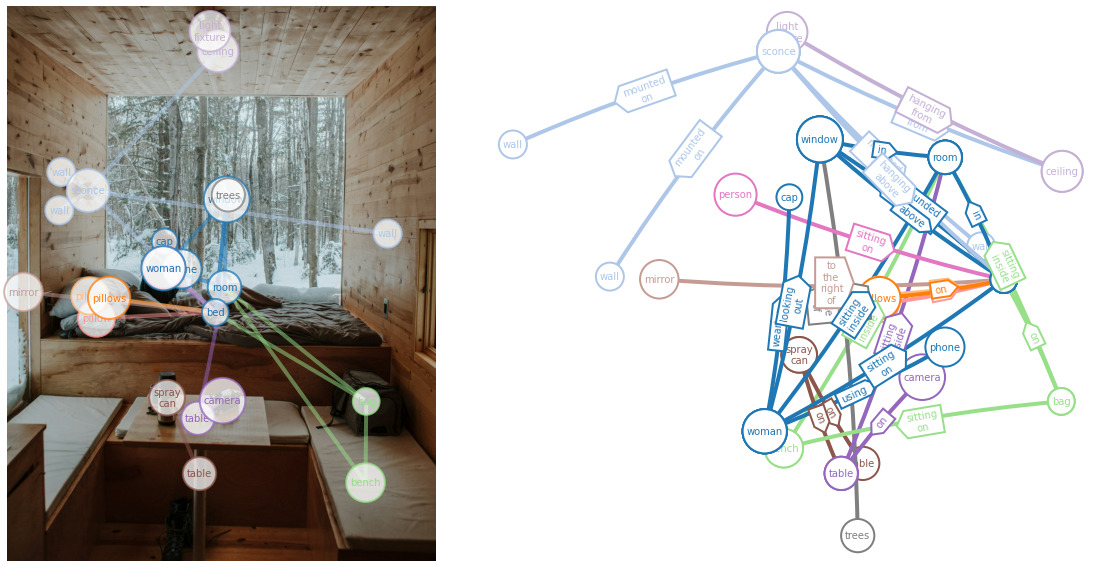}
        \caption{Photo by Nachelle Nocom on \href{https://unsplash.com/photos/woman-sitting-on-bed-watching-by-the-window-during-winter-51adhgg5KkE}{Unsplash}.}
    \end{subfigure}
    \caption{Additional graph visualizations on unseen images.}
    \label{fig:additional_graph_visualizations}
\end{figure}

\Cref{fig:additional_graph_visualizations} illustrates the ability of SG-ViT to generate entire scene graphs on novel images.
Each image shows all relationships with a score above 0.06, using the object categories and predicates from the full Visual Genome dataset to query the model.
Nodes on the left are drawn at the center of the corresponding bounding box.
Relationship predicates are shown in the graph visualization on the right.
These visualizations are intended for qualitative assessment. For downstream use of the scene graph, further post-processing, e.g. non-maximum suppression, may be applied.

\end{document}